\documentclass{article}
\pdfoutput=1

\usepackage{style}

\usepackage{graphics}
\usepackage{natbib}

\input{math_commands.sty}
\usepackage[capitalize]{cleveref}
\crefname{lemma}{lemma}{lemmas}
\Crefname{lemma}{Lemma}{Lemmas}
\crefname{thm}{theorem}{theorems}
\Crefname{corollary}{Corollary}{Corollaries}

\title{Contrastive and Non-Contrastive Self-Supervised Learning Recover Global and Local Spectral Embedding Methods}

\author{%
  Randall Balestriero\\
  Meta AI Research\\
  NYC, USA\\
  \texttt{rbalestriero@fb.com} \\
  \And
  Yann LeCun \\
  Meta AI Research \& NYU\\
  NYC, USA\\
  \texttt{ylecun@fb.com} \\
}

\begin{document}

\maketitle

\begin{abstract}
  Self-Supervised Learning (SSL) surmises that inputs and pairwise positive relationships are enough to learn meaningful representations. Although SSL has recently reached a milestone: outperforming supervised methods in many modalities\dots the theoretical foundations are limited, method-specific, and fail to provide principled design guidelines to practitioners. In this paper, we propose a unifying framework under the helm of spectral manifold learning to address those limitations. Through the course of this study, we will rigorously demonstrate that VICReg, SimCLR, BarlowTwins et al. correspond to eponymous spectral methods such as Laplacian Eigenmaps, Multidimensional Scaling et al.
  This unification will then allow us to obtain (i) the closed-form optimal representation for each method, (ii) the closed-form optimal network parameters in the linear regime for each method, (iii) the impact of the pairwise relations used during training on each of those quantities and on downstream task performances, and most importantly, (iv) the first theoretical bridge between contrastive and non-contrastive methods towards global and local spectral embedding methods respectively, hinting at the benefits and limitations of each. For example, (i) if the pairwise relation is aligned with the downstream task, any SSL method can be employed successfully and will recover the supervised method, but in the low data regime, VICReg's invariance hyper-parameter should be high; (ii) if the pairwise relation is misaligned with the downstream task, VICReg with small invariance hyper-parameter should be preferred over SimCLR or BarlowTwins.
\end{abstract}

\section{Introduction}

Self-Supervised Learning (SSL) is one of the most promising method to learn data representations that generalize across downstream tasks. SSL places itself in-between supervised and unsupervised learning as it does not require labels but does require knowledge of what makes some samples semantically close to others. Hence, where unsupervised learning relies on a collection of inputs $(\mX)$, and supervised learning relies on inputs and outputs $(\mX,\mY)$, SSL relies on inputs and inter-sample relations $(\mX,\mG)$ that indicate semantic similarity akin to weak-supervision used in metric learning \citep{xing2002distance}.
The latter matrix $\mG$ is often constructed by augmenting $\mX$ through data-augmentations known to preserve input semantics \citep{kanazawa2016warpnet,novotny2018self,gidaris2018unsupervised} e.g. horizontal flip for an image, although recent methods have went away from DA by using videos from which consecutive frames can be seen as semantically equivalent \citep{sermanet2018time,kim2019self,xu2019self}.

Although SSL originated decades ago \citep{bromley1993signature}, recent advances have pushed SSL performances beyond expectations \citep{chen2020simple,misra2020self,caron2021emerging}. Due to those rapid empirical advances, an urgent need for a principled theoretical understanding of those methods has emerged e.g. to understand how well the learned transformation transfer to different downstream tasks \citep{goyal2019scaling,ericsson2021well}. Studies looking to provide a more fundamental and principled understanding of SSL mostly take one of the three following approaches: (i) studying the training dynamics and optimization landscapes in a linear network regime e.g. validating some empirically found tricks as necessary conditions for stable gradient dynamics \cite{wang2021towards,tian2021understanding,wen2021toward,pokle2022contrasting,tian2022deep}, (ii) studying the role of individual SSL components separately e.g. the projector and predictor networks \cite{hua2021feature,jing2021understanding,bordes2021high,tosh2021contrastive}, or (iii) developing novel SSL criteria that often combine multiple interpretable objectives that a SSL model must fulfill \citep{arora2019theoretical,nazi2019generalized,wang2020understanding,shi2020run,zbontar2021barlow,bardes2021vicreg}. While those branches have led to novel understandings and even stem novel SSL methods, some fundamental questions remain open.

More recently and particularly relevant to our study, \citet{qiu2018network} unified skip-gram/word2vec methods under the helm of matrix factorization. Without much effort, those results could be used to unify pretext-task SSL learning \citep{baevski2020wav2vec,bao2021beit,he2021masked} (our focus is on joint-embedding SSL). Closer to our topic, \citet{haochen2021provable,haochen2022beyond} developed a study of SimCLR \citep{chen2020simple} by proposing a modified objective coined as Spectral Contrastive Loss based on graph representation of pairwise similarities. A perhaps more direct connection between video-based SSL and spectral methods (Laplacian Eigenmap in particular) can be found by combining the Spectral Inference Networks (SIN) of \citet{pfau2018spectral} that generalize Slow Feature Analysis (SFA) to arbitrary linear operators, and the known relationship between SFA and Laplacian Eigenmap \citep{sprekeler2011relation}. Although SIN was not framed within an SSL viewpoint e.g. SIN has a well-posed learned coordinate system while SSL methods coordinates can be arbitrarily rotated; those studies paved our way forward as {\em we propose in this paper a broad analysis that unify most existing SSL methods as variants of known spectral embedding methods, allowing us for the first time to provide provable design guildelines to practitioners in their choice of architecture and methods}. We summarize our unification results in \cref{fig:summary}. The instrumental results we obtain allow us to answer some long-standing questions such as:

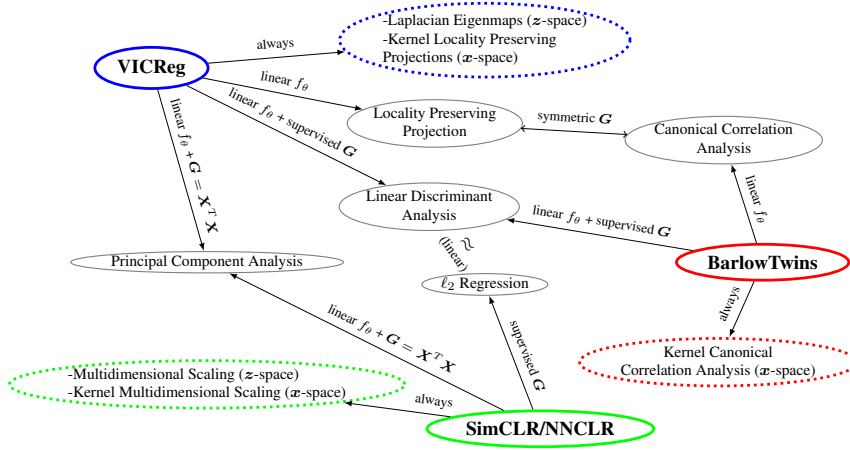
\begin{figure}[t!]
    \centering
    \begin{minipage}{0.75\linewidth}
    \resizebox{\linewidth}{!}{
\begin{tikzpicture}
    
\node[ellipse,draw=blue,line width=0.5mm,inner sep=3pt,align=center] at (-6,-0.5) (vicreg) {\bf VICReg};

\node[ellipse, draw=red,line width=0.5mm,inner sep=3pt,align=center] at (5,-4) (barlow) {\bf BarlowTwins};

\node[ellipse, draw=green,line width=0.5mm,inner sep=3pt,align=center] at (1,-7) (simclr) {\bf SimCLR/NNCLR};

\node[ellipse, dotted, draw=blue,line width=0.5mm,inner sep=15pt,align=center,inner sep=0.01cm] at (0,0) (LLE) {\begin{varwidth}{\linewidth}\footnotesize
-Laplacian Eigenmaps ($\vz$-space)\\
-Kernel Locality Preserving\\
Projections ($\vx$-space)\end{varwidth}};

\node[ellipse, dotted, draw=red,line width=0.5mm,inner sep=15pt,align=center,inner sep=0.01cm] at (4.2,-5.8) (DCA) {\begin{varwidth}{\linewidth}\footnotesize\centering Kernel Canonical\\Correlation Analysis ($\vx$-space)\end{varwidth}};

\node[ellipse, draw=gray,line width=0.2mm,inner sep=15pt,align=center,inner sep=0.01cm] at (4.3,-1.8)  (CCA) {\begin{varwidth}{\linewidth}\footnotesize\centering Canonical Correlation\\Analysis\end{varwidth}};

\node[ellipse, draw=gray,line width=0.2mm,inner sep=15pt,align=center,inner sep=0.01cm] at (-0.9,-1.5)  (LLP) {\begin{varwidth}{\linewidth}\footnotesize\centering Locality Preserving\\Projection\end{varwidth}};

\node[ellipse, draw=gray,line width=0.2mm,inner sep=15pt,align=center,inner sep=0.01cm] at (-1.,-3.)  (LDA) {\begin{varwidth}{\linewidth}\footnotesize\centering Linear Discriminant\\Analysis\end{varwidth}};

\node[ellipse, draw=gray,line width=0.2mm,inner sep=15pt,align=center,inner sep=0.01cm] at (-0.,-4.4)  (LR) {\begin{varwidth}{\linewidth}\footnotesize\centering $\ell_2$ Regression\end{varwidth}};

\node[ellipse, draw=gray,line width=0.2mm,inner sep=15pt,align=center,inner sep=0.01cm] at (-5,-4)  (PCA) {\begin{varwidth}{\linewidth}\footnotesize
Principal Component Analysis\end{varwidth}};

\node[ellipse, dotted, draw=green,line width=0.5mm,inner sep=15pt,align=center,inner sep=0.01cm] at (-5,-6.2)  (ISOMAP) {\begin{varwidth}{\linewidth}\footnotesize
-Multidimensional Scaling ($\vz$-space)\\
-Kernel Multidimensional Scaling ($\vx$-space)\end{varwidth}};

\path [draw,-latex] (simclr) -- node [pos=0.17,above,sloped] {\scriptsize always} (ISOMAP);
\path [draw,-latex] (simclr) -- node [above,sloped,pos=0.43] {\scriptsize linear $f_{\theta}$ + $\mG=\mX^T\mX$} (PCA);
\path [draw,-latex] (simclr) -- node [above,sloped,pos=0.43] {\scriptsize supervised $\mG$} (LR);
\path [draw,-latex] (vicreg) -- node [midway,above,sloped] {\scriptsize linear $f_{\theta}$ + $\mG=\mX^T\mX$} (PCA);
\path [draw,-latex] (vicreg) -- node [midway,above,sloped] {\scriptsize linear $f_{\theta}$} (LLP);
\path [draw,-latex] (vicreg) -- node [midway,above,sloped] {\scriptsize linear $f_{\theta}$ + supervised $\mG$} (LDA);
\path [draw,-latex] (vicreg) -- node [midway,above,sloped] {\scriptsize always} (LLE);
\path [draw,-latex] (barlow) -- node [midway,above,sloped] {\scriptsize linear $f_{\theta}$ + supervised $\mG$} (LDA);
\path [draw,-latex] (barlow) -- node [midway,above,sloped] {\scriptsize linear $f_{\theta}$} (CCA);
\path [draw,-latex] (barlow) -- node [midway,above,sloped] {\scriptsize always} (DCA);
\path (LR) -- node [pos=0.56,sloped] {\centering\begin{varwidth}{\linewidth}\centering\scriptsize
{\normalsize$\approx$}\\
 (linear)\end{varwidth}} (LDA);
\path [draw,<->] (LLP) -- node [midway,above,sloped] {\scriptsize symmetric $\mG$} (CCA);
\end{tikzpicture}
}
\end{minipage}
\begin{minipage}{0.24\linewidth}
\caption{\small Summary of our unification of SSL methods to known {\em local} and {\em global} spectral embedding methods. In doing so, we are able to find the exact settings for which different methods provably become identical. In short, all are concerned in preserving the left-singular vectors of the similarity matrix $\mG$ (see \cref{fig:G}) in the representation $\mZ$.
    }
    \label{fig:summary}
\end{minipage}
\end{figure}

{\em Are the numerous flavors of SSL methods e.g. contrastive and non-contrastive learning different representations?}
We demonstrate through this study that all SSL methods' optimal learn a representation $\mZ$ whose top left singular vectors align with the ones of $\mG$, and that none of the SSL methods constrain the right-singular vectors of $\mZ$.

{\em Can we guarantee that minimizing a SSL loss produces a representation that is optimal to solve a downstream task?} Yes, \cref{sec:same_coin} demonstrates that a representation learned from $(\mX,\mG)$ with any SSL loss is guaranteed to solve any downstream task $(\mX,\mY)$ as long as the left spectrum of $\mG$ and $\mY$ are aligned.

{\em Are there fundamental differences between contrastive and non-contrastive methods e.g. when $\mG$ is unknown?} We demonstrate that the optimal VICReg representation can be made full-rank while learning from $\mG$ by carefully selecting the loss hyperparameters (\cref{thm:VICReg_optimal,fig:VICReg_landscape2}), while SimCLR and BarlowTwins strictly enforce $\rank(\mZ)=\rank(\mG)$ (\cref{fig:simclr,fig:BT} respectively), hinting at a possible advantage of VICReg when $\mG$ is misspecified (\cref{sec:not_equal}).

{\em How does connecting SSL methods to spectral embedding methods improve our understanding and guide the design of novel SSL frameworks?} We demonstrate that contrastive and non-contrastive SSL corresponds to global and local spectral embedding methods respectively (\cref{sec:vicreg_le,sec:simclr_isomap}). From that, we easily identify the best use-cases for each of them e.g. contrastive methods aim to be metric preserving and shine with low-dimensional or high-dimensional but linear manifolds while non-contrastive shine with locally linear but globally nonlinear manifolds.

The natural symbiosis between SSL and spectral methods formulations that will be discovered throughout this study not only enables us to answer the above challenging questions but also provides practical guidelines to practitioners. One emblematic example is offered in \cref{sec:vicreg_linear,sec:barlowtwins_linear} where we obtain the analytical network parameters of SSL methods in the linear regime; another example is offered in \cref{sec:simclr_isomap} where novel and interpretable variations of SimCLR are obtained from first principles; or even in \cref{sec:not_equal} where we identify the situations for which different SSL methods should be preferred. 
\\
We summarize our contributions below:
\begin{enumerate}
    \item {\bf Closed-form optimal representation for SSL losses.}~The DN representation $\mZ$ of inputs $\mX$ learned by minimizing {\em any} SSL loss given a sample relation matrix $\mG$ is obtained in closed-form, shedding light to many spectral properties of those representation e.g. SSL only constrains the left singular vectors and singular values of $\mZ$ to align with the ones of $\mG$ (\cref{sec:vicreg_general,sec:simclr_first,sec:BT_first} for VICReg, SimCLR and BarlowTwins).
    \item {\bf Closed-form optimal network parameters for SSL losses with linear networks.}~The linear representation $\mZ=\mX\mW+\vb$ parameters obtained by minimizing {\em any} SSL loss given a sample relation matrix $\mG$ are obtained in closed-form, providing insights into the type of input statistics that a network parameters focus on to produce the optimal input representation (\cref{sec:vicreg_linear,sec:barlowtwins_linear} for VICReg and BarlowTwins).
    \item {\bf Exact equivalence between SSL and spectral embedding methods.}~SSL methods employ diverse criterion that can be tied to eponymous spectral analysis methods both when employing a nonlinear DN as Laplacian Eigenmaps (VICReg, \cref{sec:vicreg_le}), ISOMAP (SimCLR/NNCLR, \cref{sec:simclr_isomap}), Canonical Correlation Analysis (BarlowTwins, \cref{sec:barlowtwins_linear}) and when employing a linear network as Locality Preserving Projection (VICReg, \cref{sec:vicreg_linear}), Cannonical Correlation Analysis (BarlowTwins, \cref{sec:barlowtwins_linear}), and Linear Discriminant Analysis for both VICReg and BarlowTwins.
    
    \item {\bf Optimality conditions of SSL representations on downstream tasks ($\mY$).}~When the correct data relation matrix is given i.e. with left singular vector associated to nonzero singular values that span the space of left singular vectors of the target matrix, then perfect minimization of any of those SSL losses will provide an optimal representation (\cref{sec:same_coin}) which ---up to a rotation of its right singular vectors--- is identical to the one learned in a supervised setting (\cref{sec:MSE}) a necessary and sufficient condition to perfectly solve a task at hand with a linear probe (\cref{sec:conditions}).
\end{enumerate}

We carefully prove each statement of this study in \cref{proof:section}. To ensure clarity of our statements and results, we also provide code excerpts throughout the manuscript.
\section{Notations and Background on Self-Supervised Learning}
\label{sec:background}

We provide in this section a brief reminder of the main Self-Supervised Learning (SSL) methods, their associated losses, and the common notations that we will rely on for the remaining of the study.

{\bf Dataset, Embedding and Relation Matrix Notations.}~Regardless of the loss and method employed, SSL relies on having access to a set of observations i.e. input samples $\mX\triangleq [\vx_1, \dots, \vx_{N}]^T \in \mathbb{R}^{N \times D}$ and a known pairwise positive relation between those samples e.g. in the form of a {\em symmetric} matrix $\mG \in \{0,1\}^{N \times N}$ where $(\mG)_{i,j}=1$ iff samples $\vx_i$ and $\vx_j$ are known to be semantically related, and with $0$ in the diagonal. Commonly, one is only given a dataset $\mX'\in\mathbb{R}^{N' \times D'}$ where commonly $D=D'$, and artificially constructs $\mX,\mG$ from augmentations of $\mX'$ e.g. rotated, noisy versions of the original samples and turning the corresponding entries of $\mG$ to be positive for the samples that have been augmented form the same original sample. The positive samples are often denoted as {\em views}, and in the situation where only $\mX'$ is given, $\mX$ is often formed as 
\begin{align}
    \mX = [\View_1(\mX')^T,\dots,\View_V(\mX')^T]^T,\label{eq:Xprime}
\end{align}
where a row $(\View_v(\mX'))_{n,.}$ is viewed as similar to the same row of $(\View_{v'}(\mX'))_{n,.},\forall v \not = v'$. Commonly, one employs $V=2$ i.e. only positive {\em pairs} are used. If one desires to duplicate a sample multiple times to obtain multiple positive pairs from the same original sample $\vx_n'$, it is done by duplicating that sample within $\mX'$ prior applying \cref{eq:Xprime}. The $\View_c(.),c=1,\dots,C$ operator is a sample-wise transformation e.g. adding white noise, masking, and the likes i.e. the same row of different view-matrix are different transformations of the same original sample from $\mX'$. In the case of \cref{eq:Xprime}, $\mG$ ---which will be of shape $(VN' \times VN')$--- can be easily formed via 
\begin{lstlisting}[language=Python,escapechar=\%]
G = torch.zeros(Np * V, Np * V) #  %\lsComment{\scriptsize $\mX'\in \mathbb{R}^{N' \times D'}$}%, V is the number of views
i = torch.arange(0, Np * V).repeat_interleave(V - 1) # row indices
j = (i + torch.arange(1, V).repeat(Np * V) * Np).remainder(Np * V) # column indices
G[i,j] = 1 # unweighted graph connecting the rows of %\lsComment{\scriptsize $\View_1(\mX'), \dots, \View_V(\mX')$}%
\end{lstlisting}
where a sparse matrix could be used for $\mG$ to improve efficiency, examples are provided in \cref{fig:G}. 

Lastly, we will also denote by $\mZ\in\mathbb{R}^{N \times K}$ the matrix of feature maps  or {\em embeddings} obtained from a model $f_{\theta}$ ---commonly a Deep Network--- such as $\mZ\triangleq [f_{\theta}(\vx_1), \dots, f_{\theta}(\vx_{N})]^T$

{\bf VICReg.}~With the above notations out of the way, we can introduce the {\em VICReg} loss \citet{bardes2021vicreg} which is a function of $\mX$ and $\mG$ ---although $\mG$ was not explicitly employed originally--- as
\begin{align}
\mathcal{L}_{\rm vic}\hspace{-0.05cm}=&\alpha\hspace{-0.05cm} \sum_{k=1}^{K}\max\hspace{-0.05cm}\left(0,1\hspace{-0.05cm}-\hspace{-0.05cm}\sqrt{\Cov(\mZ)_{k,k}}\right)\hspace{-0.08cm}+\hspace{-0.05cm} \beta\hspace{-0.1cm} \sum_{j=1,j\not = k}^{K}\hspace{-0.15cm}\Cov(\mZ)^2_{k,j}\hspace{-0.05cm}+\hspace{-0.05cm}\frac{\gamma}{N} \sum_{i=1}^{N}\sum_{j=1}^{N}(\mG)_{i,j}\|\mZ_{i,.}-\mZ_{j,.}\|_2^2.\label{eq:VICReg}
\end{align}
The VICReg loss has a computational complexity of $\mathcal{O}(NK^2+PNK)$ with $P$ the average number of positive samples i.e. number of nonzeros elements in each row of $\mG$. Since $P$ is often small ($P \ll K$), the computational cost is dominated by the covariance matrix, $\mathcal{O}(NK^2)$. The implementation of \cref{eq:VICReg} is straightforward (details in the proof of \cref{thm:VICReg_optimal}) as
\begin{lstlisting}[language=Python,escapechar=\%]
C = torch.cov(Z.t()) # %\lsComment{\scriptsize $\mZ \in \mathbb{R}^{N \times K}$}%
var_loss = K - torch.diag(C).clamp(eps, 1).sqrt().sum() # %\lsComment{$\epsilon$}% avoids inf. sqrt gradient at 0
i,j = G.nonzero(as_tuple=True) # with G as in Fig. %\lsComment{\ref{fig:G}}%, i and j are vectors of indices
inv_loss =(Z[i]-Z[j]).square().sum(1).inner(G[i,j])/N # pairwise %\lsComment{$\ell_2$}% weighted by %\lsComment{$\mG_{i,j}$}%
cov_loss = 2 * torch.triu(C, diagonal=1).square().sum() # 2x upper triangular part
loss = alpha * var_loss + beta * cov_loss + gamma * inv_loss # equals to %\lsComment{(\ref{eq:VICReg})} with $\alpha,\beta,\gamma$%
\end{lstlisting}

{\bf SimCLR.}~The SimCLR loss \citep{chen2020simple} consists in two steps. First, it produces an estimate $\widehat{\mG}(\mZ)$ of the relation matrix $\mG$ from the embeddings $\mZ$, generally by using the cosine similarity ($\CosSim$) as in
\begin{align}
    (\widehat{\mG}(\mZ))_{i,j}=\frac{e^{\CosSim(\vz_i,\vz_j)/\tau}}{\sum_{k=1,k\not = i}^{N}e^{\CosSim(\vz_i,\vz_k)/\tau}}\Indic_{\{i=j\}},\label{eq:simCLR}
\end{align}
with $\tau>0$ a temperature parameter. Notice that $\widehat{\mG}(\mZ)$ is a right-stochastic matrix i.e. $\widehat{\mG}(\mZ)\mathbf{1}=\mathbf{1}$. Then, SimCLR encourages the elements of $\widehat{\mG}(\mZ)$ and $\mG$ to match. The most popular solution to achieve that is to leverage the infoNCE loss given by
\begin{align}
    \mathcal{L}_{\rm simCLR}=\underbrace{-\sum_{i=1}^{N}\sum_{j=1}^{N}(\mG)_{i,j}\log(\widehat{\mG}(\mZ))_{i,j}}_{\text{cross-entropy between matrices}},\label{eq:simclr}
\end{align}
where $\mG$ is assumed to be a right-stochastic matrix as well. If not, a simple renormalization can be applied before employing the infoNCE loss.
The only difference between SimCLR and its variants e.g. NNCLR \citep{dwibedi2021little} lies in the construction of $\mG$ when only given $\mX'$ (the input samples without any augmentation applied, recall the beginning of \cref{sec:background}). As opposed to VICReg, the SimCLR loss has computational complexity of $\mathcal{O}(N^2)$ as it requires to compute all the pairwise similarities in \cref{eq:simCLR}. Nevertheless, \cref{eq:simclr} can be easily computed as follows (here using the $\CosSim$ again)
\begin{lstlisting}[language=Python,escapechar=\%]
Z_renorm = torch.nn.functional.normalize(Z, dim=1) # %\lsComment{\scriptsize $\mZ \in \mathbb{R}^{N \times K}$}%
cosim = Z_renorm @ Z_renorm.t() / tau # N x N matrix, tau is the temperature %\lsComment{$\tau$}% from %\lsComment{(\ref{eq:simCLR})}%
off_diag = cosim[~I.bool()].reshape(N, N-1) # for denom. of %\lsComment{(\ref{eq:simCLR})}%, I = torch.eye(N, N)
loss = (G * (torch.logsumexp(off_diag, dim=1, keepdim=True) - cosim)).sum() # gives %\lsComment{(\ref{eq:simclr})}%
\end{lstlisting}

{\bf BarlowTwins.}~Lastly, {\em BarlowTwins} \citep{zbontar2021barlow} proposes yet a slightly different approach based on observing two views of a dataset $\mX'$ denoted as $\mX_{\rm left}$ and $\mX_{\rm right}$ and with corresponding embeddings $\mZ_{\rm left}$ and $\mZ_{\rm right}$. The same row of those left/right matrices are the positive pairs. 
Denoting by $\mC$ the $K \times K$ cross-correlation matrix between $\mZ_{\rm left}$ and $\mZ_{\rm right}$, we obtain the loss
\begin{align}
    \mathcal{L}_{\rm BT}=\sum_{k=1}^{K}((\mC)_{k,k}-1)^2+\alpha \sum_{k' \not = k}(\mC)_{k,k}^2,\;\alpha > 0.\label{eq:BT}
\end{align}
Notice that $(\mC)_{i,j}$ falls back to measuring the cosine similarity between the $i^{\rm th}$ column of $\mZ_{\rm left}$ and the $j^{\rm th}$ column of $\mZ_{\rm right}$ i.e. $(\mC)_{i,j}=\frac{\langle (\mZ_{\rm left})_{.,i},(\mZ_{\rm right})_{.,j}\rangle}{\|(\mZ_{\rm left})_{.,i}\|_2\|(\mZ_{\rm right})_{.,j}\|_2+\eps}$ where one commonly adds an additional constant $\epsilon$ for numerical stability. The computational complexity of \cref{eq:BT} falls back to $\mathcal{O}(NK^2)$ and its computation is easily performed as
\begin{lstlisting}[language=Python,escapechar=\%]
Z_lrenorm = torch.nn.functional.normalize(Z_left, dim=0, eps=eps) # includes %\lsComment{$\epsilon$}%
Z_rrenorm = torch.nn.functional.normalize(Z_right, dim=0, eps=eps) # includes %\lsComment{$\epsilon$}%
C = Z_lrenorm.t() @ Z_rrenorm # cross-correlation (K x K matrix)
loss = (C.diag() - 1).square().sum() + alpha*C[~I.bool()].square().sum() # gives %\lsComment{\cref{eq:BT}}%
\end{lstlisting}
Note that if one possesses a dataset with arbitrary number of positive samples, it is still possible to recover $\mX_{\rm left},\mX_{\rm right}$ for $\mathcal{L}_{\rm BT}$ with the following simple strategy. Suppose that we have $5$ samples $\va,\vb,\vc,\vd,\ve$, and that $\va,\vb,\vc$ are related to each other, and that $\vd,\ve$ are related to each other, based on $\mG$. Then, one can simply create the two data matrices as
\begin{align}
 \mX_{\rm left}=[\va,\va,\vb,\vb,\vc,\vc,\vd,\ve]^T,\;\mX_{\rm right}=[\vb,\vc,\va,\vc,\va,\vb,\ve,\vd]^T,\label{eq:pair_form}
\end{align}
which can be easily obtained given $\mG$ and the input matrix $\mX$ via
\begin{lstlisting}[language=Python,escapechar=\%]
row_indices, col_indices = G.nonzero(as_tuple=True) # G as in Fig. %\lsComment{\ref{fig:G}}%
X_left = X[row_indices]  # equals the left-hand-side of %\lsComment{(\ref{eq:pair_form})}%
X_right = X[col_indices]  # equals the right-hand-side of %\lsComment{(\ref{eq:pair_form})}%
\end{lstlisting}

\begin{figure}[t!]
    \centering
    \begin{minipage}{0.32\linewidth}\small
    \centering
    binary classification graph\\[1.2em]
    \includegraphics[width=\linewidth]{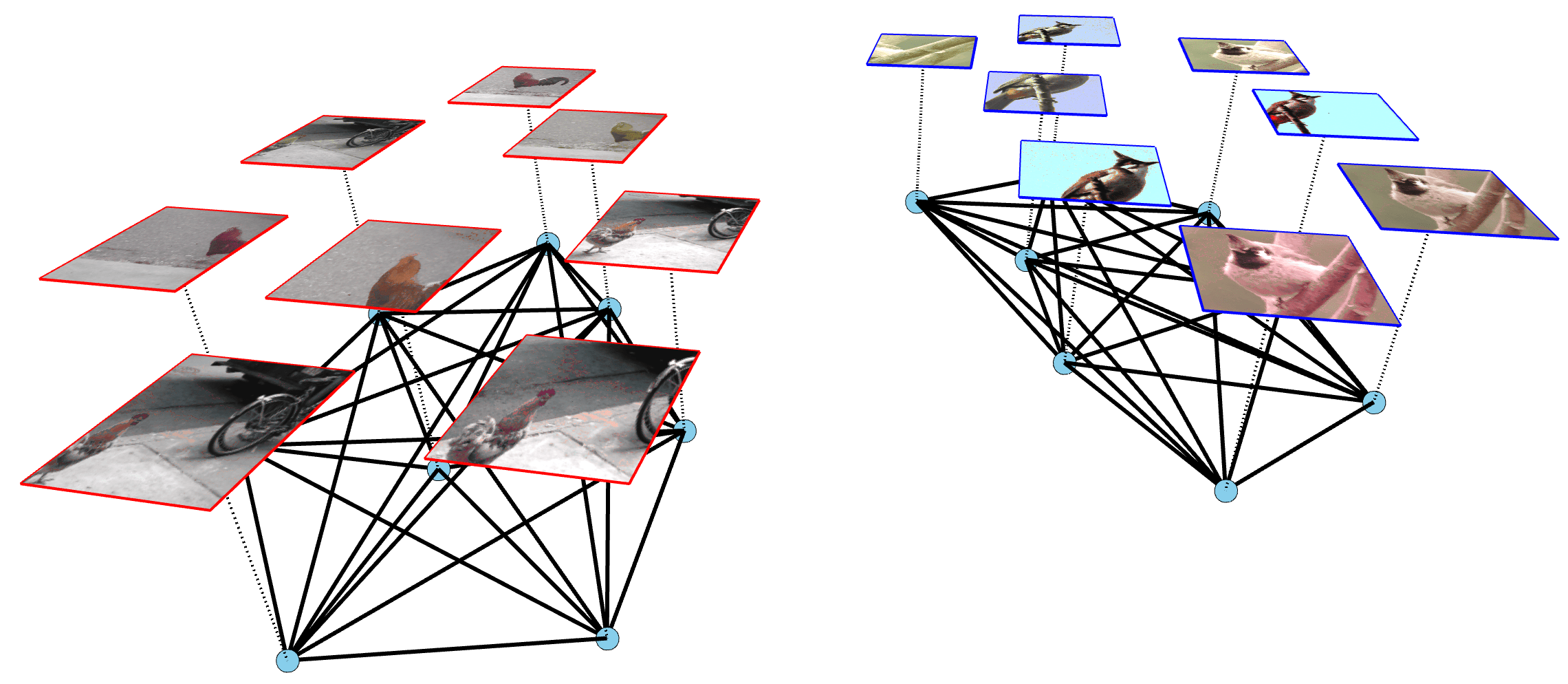}
    \end{minipage}
    \begin{minipage}{0.32\linewidth}\small
    \centering
    ``multi-crop-type'' self-supervised graph\\
    \includegraphics[width=\linewidth]{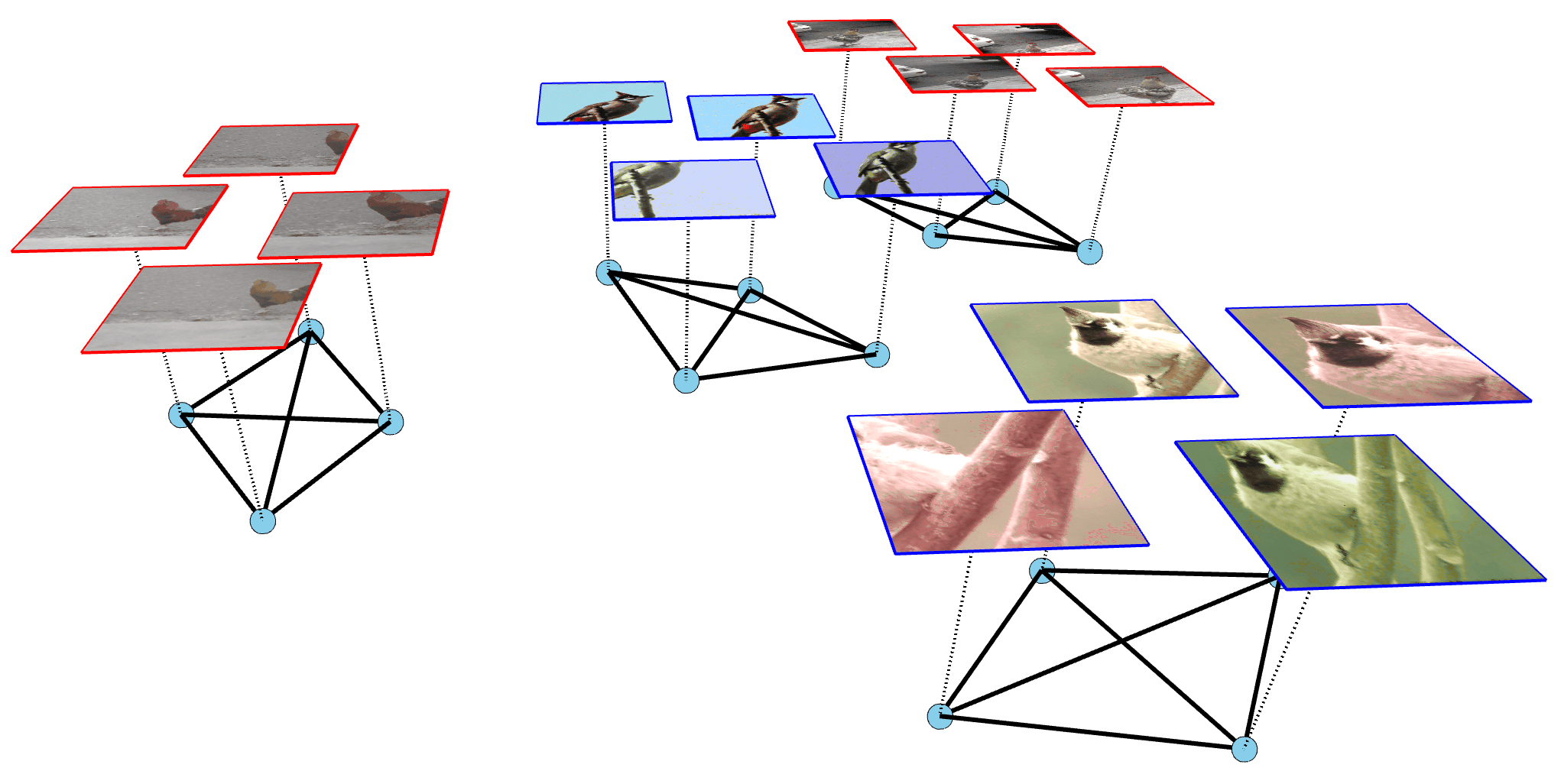}
    \end{minipage}
    \begin{minipage}{0.32\linewidth}\small
    \centering
    ``pair-type'' self-supervised graph\\
    \includegraphics[width=\linewidth]{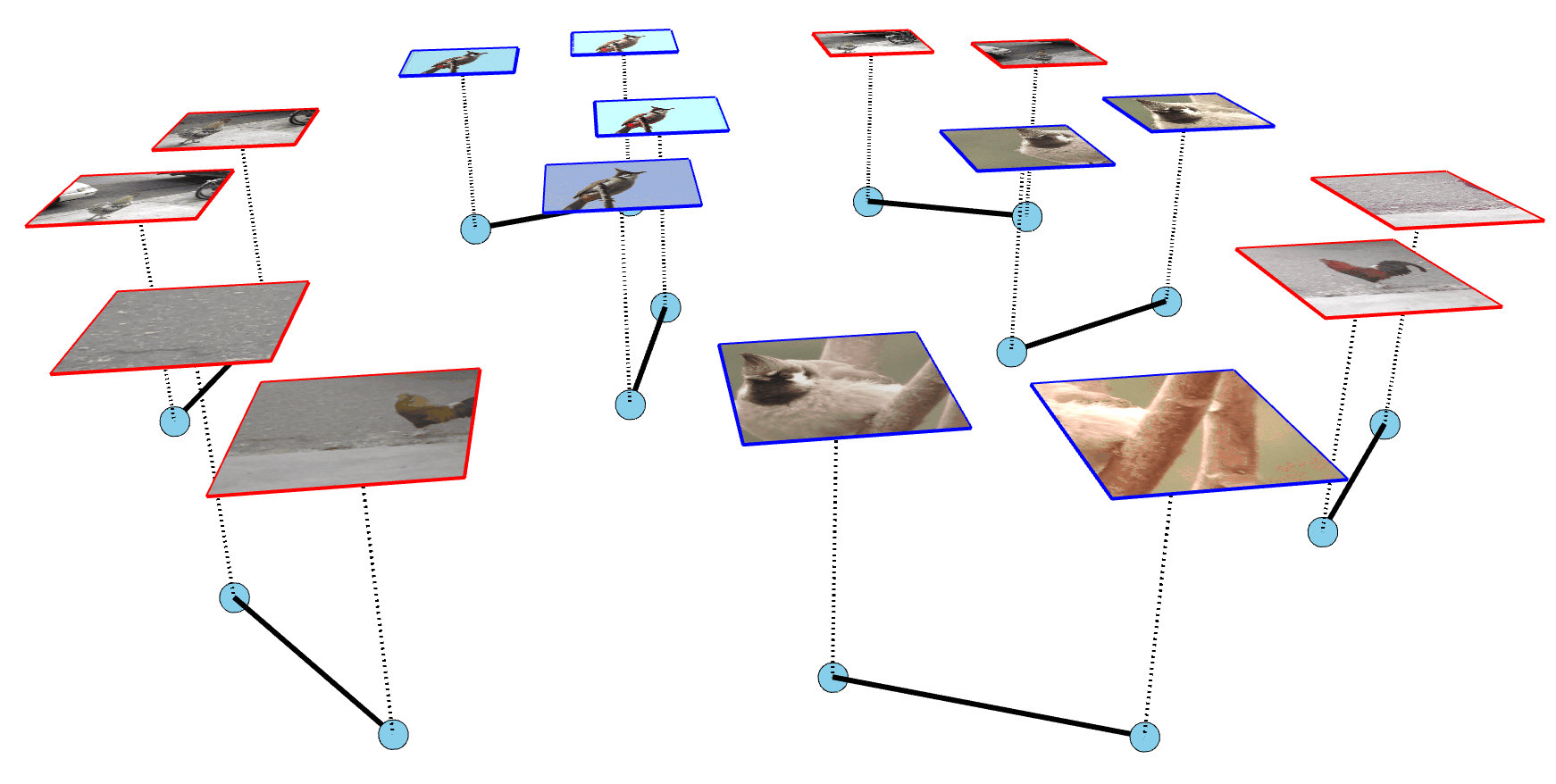}
    \end{minipage}
    \caption{\small Depiction of the ``knowledge graph'' arising from different settings: binary classification ({\bf left}), SSL with multi-crop-style relation i.e. that number of positive samples associated to each datum can be made arbitrarily large ({\bf middle}), and pair-style SSL where each sample is only associated to another single sample ({\bf right}). In this representation, each point can be equivalently seen as an abstract node ({\bf blue circlr}) or as a single signal $\vx_n,n=1,\dots,N$ (image in this case, {\bf depicted on top}). In this case, the underlying data comes from two classes depicted with red and blue image borders. A model's prediction on those signals $f_{\theta}(\vx_n)$ can be viewed as a signal on the graph (recall \cref{sec:background}). That graph is itself encoded as a symmetric, nonnegative matrix similarity matrix $\mG$, as depicted in \cref{fig:G}.}
    \label{fig:teaser}
\end{figure}

\begin{figure}[t!]
    \centering
    \begin{minipage}{0.6\linewidth}
    \begin{minipage}{0.32\linewidth}\small
    \centering
    binary classification\\(2 classes {\tiny $\mX$=$[\mX_0^T\hspace{-0.05cm},\hspace{-0.05cm}\mX_1^T]^T$})
    \end{minipage}
    \begin{minipage}{0.32\linewidth}\small
    \centering
    ``multi-crop-type'' SSL\\(N/K classes, K>2)
    \end{minipage}
    \begin{minipage}{0.32\linewidth}\small
    \centering
    ``pair-type'' SSL\\(N/2 classes)
    \end{minipage}
    \includegraphics[width=\linewidth]{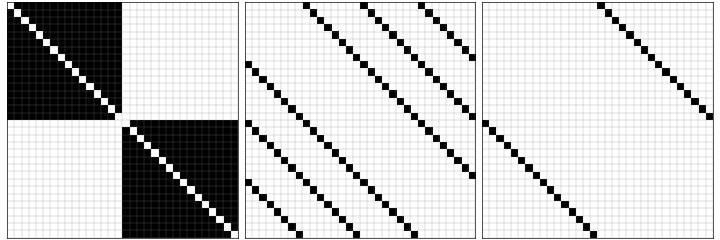}
    \end{minipage}
    \begin{minipage}{0.39\linewidth}
    \caption{\small Examples of the $N \times N$ symmetric adjacency matrices $\mG$ for the three cases of \cref{fig:teaser}. In all cases, each nonzero entry $(\mG)_{i,j}$ represents the positive relation between sample $i$ and $j$. The insight that will play a key role in our analysis is that the eigenvectors of those matrices entirely encode the similarity information between samples that SSL utilize to constrain the learned representation ($\mZ$)'s spectrum (see \cref{thm:VICReg_optimal,thm:simclr,thm:BT}). Although $\mG$ is commonly symmetric, it is often not required (see \cref{thm:BT_cca}).}
    \label{fig:G}
    \end{minipage}
\end{figure}

{\bf Linear Algebra Notations.}~
This study heavily relies on the Singular Value Decomposition (SVD) of matrices \citep{eckart1936approximation,hestenes1958inversion} e.g. $\mX=\mU_{x}\mSigma_{x}\mV^T_{x}$ that are denoted as the left singular vectors $\mU_{x} \in \mathbb{R}^{N \times N}$, the singular values $\mSigma_{x} \in \mathbb{R}^{N \times D}$ and the right singular vectors $\mV_{x} \in \mathbb{R}^{D \times D}$ of $\mX\in\mathbb{R}^{N \times D}$. We will always specify as a lower-script and lower-case the matrix that is being decomposed ($x$ in this case). It will also be convenient to only consider the left/right singular vectors whose associated singular values are $0$ that we will denote as $\overline{\mU}_{x}$ and $\overline{\mV}_{x}$ respectively. Conversely, the left/right singular vectors whose associated singular values are $>0$ will be denote as $\widehat{\mU}_{x}$ and $\widehat{\mV}_{x}$ respectively. Lastly, we will denote by $\vsigma_z$ the vector of singular values such as $\mSigma_{z}=\diag(\vsigma)$ and without loss of generality and unless otherwise stated, this will always be in descending order.

Our goal in the following sections (\cref{sec:vicreg} for VICReg, \cref{sec:simclr} for SimCLR/NNCLR, and \cref{sec:BT} for BarlowTwins) will be to find the optimal representations $\mZ$ of $\mX$ whilst tying those methods to their spectral embedding counterpart. Three surprising facts will emerge: (i) all existing methods recover exactly some flavors of famous spectral method, (ii) the spectral properties of the optimal representation $\mZ$ of $\mX$ can be obtained in closed-form, and (iii) from those properties, necessary and sufficient conditions can be obtained to bounds the downstream task error of those optimal SSL representations (\cref{sec:optimal}).
\section[VICReg]{VICReg Minimizes the Dirichlet Energy to Produce Smooth Signals on the Graph $\mG$ While Preventing Dimensional Collapse}
\label{sec:vicreg}

We will first demonstrate  in \cref{sec:vicreg_general} that the optimal VICReg representation can be obtained in closed form (\cref{thm:VICReg_optimal}) and that turning the VICReg optimization as a constrained problem recovers (Kernel) Laplacian Eigenmaps, an eponymous spectral embedding method \cref{sec:vicreg_le}. We then consider that same constrained problem but under a linear network regime for which we obtain the closed-form optimal network's parameters (\cref{thm:vicreg_linear}) and in which case, VICReg recovers Locality Preserving Projections and Linear Discriminant Analysis (\cref{sec:vicreg_linear}).

\subsection{Closed-Form Optimal Representation for VICReg}
\label{sec:vicreg_general}

The first goal of this section is to build up some insights into VICReg by demonstrating how the invariance term corresponds to the Dirichlet energy of the signal $\mZ$ on the graph $\mG$. Then, replacing the variance hinge loss at $1$ with the squared loss at $1$ as in $\sum_{k=1}^{K}\left(1-\Cov(\mZ)_{k,k}\right)^2$ ---notice that minimizing the latter implies minimizing the former--- we obtain the closed-form optimal representation $\mZ^*$ of $\mX$ which turns out to be a function only of $\mG$ and the loss' hyper-parameters.

{\bf From invariance to trace minimization.}~As a first step, we propose to better understand the impact of VICReg's invariance term (recall \cref{eq:VICReg}) onto the left-singular vectors of the representation $\mZ$. Recalling from \cref{sec:background} that we denote the SVD of $\mZ$ by $\mU_{\mZ}\mSigma_{\mZ}\mV^T_{\mZ}$ we obtain the following
\begin{align*}
    \underbrace{(\mG)_{i,j}\|(\mZ)_{i,.}-(\mZ)_{j,.} \|_2^2}_{\text{$(i,j)$ invariance term in VICReg}} =  (\mG)_{i,j}\|\mSigma_z^T ((\mU_{\mZ})_{i,.}-(\mU_{\mZ})_{j,.}) \|_2^2\propto\underbrace{ (\mG)_{i,j}\|(\widehat{\mU}_{\mZ})_{i,.}-(\widehat{\mU}_{\mZ})_{j,.} \|_2^2}_{\mathclap{\text{acts upon the rows of the left singular vector of $\mZ$}}},
\end{align*}
where we recall that $\widehat{\mU}_{\mZ}$ are the left-singular vectors of $\mZ$ associated to nonzero singular values, and $(\mM)_{i,.}$ extracts the $i^{\rm th}$ row viewed as a column vector, for any matrix $\mM$. Hence, VICReg effectively minimizes the pairwise distance between the $i^{\rm th}$ and $j^{\rm th}$ rows of $\widehat{\mU}_{Z}$ whenever $(\mG)_{i,j}>0$. This connection can be made more precise by rewriting the invariance loss of VICReg as the energy of the signal $\mZ$ on the graph $\mG$ \citep{von2007tutorial} since we have (derivations in \cref{proof:invariance_trace})
\begin{align}
    \underbrace{\sum_{i}\sum_{j}(\mG)_{i,j}\|(\mZ)_{i,.}-(\mZ)_{j,.}\|_2^2}_{\text{invariance term in VICReg}}=2\underbrace{\Tr\left(\mZ^T\mL\mZ\right)}_{\mathclap{\text{Dirichlet energy of $\mZ$ on $\mG$}}},\label{eq:VICReg_trace}
\end{align}
where $\mL$ is the graph Laplacian matrix $\mL=\mD-\mG$ with $\mD$ the diagonal degree matrix of $\mG$ i.e. $(\mD)_{i,j}=\sum_{j}(\mG)_{i,j}$ and $(\mD)_{i,j}=0, \forall i \not = j$. From \cref{eq:VICReg_trace} it is clear that the invariance term depends on the matching between the left singular vectors of $\mZ$ and the eigenvectors of $\mL$. Hence, {\em non-contrastive learning aims at producing non-degenerate signals $\mZ$ that are smooth on $\mG$}.

{\bf Analytical optimal representation.}~To gain further insights into VICReg, we ought to obtain the analytical form of the optimal representation $\mZ^*$ minimizing \cref{eq:VICReg} ---although this optimum is not unique e.g. adding a constant entry to each column of $\mZ$ does not change the loss value (details in \cref{proof:not_unique}). To that end, we will need to work with a slightly friendlier variance term i.e. we replace the hinge loss at $1$ with the squared loss centered at $1$ as in $\sum_{k=1}^{K}\left(1-\Cov(\mZ)_{k,k}\right)^2$. We can now obtain the following characterization of $\mZ^*$ as a function of the spectral decomposition of the matrix that combines two Laplacian matrices (details in \cref{proof:VICReg_optimal}). The first, (left of \cref{eq:VICReg_optimal}) comes form the variance+covariance term is the Laplacian of a complete graph i.e. where each node/sample is connected all others, and the second (right of \cref{eq:VICReg_optimal}) is the one of the graph described by $\mG$ as in
\begin{align}
    \underbrace{\mI-\frac{1}{N}\mathbf{1}\mathbf{1}^T}_{\mathclap{\text{Laplacian of a complete graph}}}-\frac{\gamma}{\alpha} \overbrace{(\mD-\mG)}^{\mathclap{\text{Laplacian of the SSL/sup. graph}}}=\mP_{\alpha,\gamma}\diag(\vlambda_{\alpha,\gamma}) \mP^T_{\alpha,\gamma},\label{eq:VICReg_optimal}
\end{align}
where the eigenvalues/eigenvectors are in descending orders. Combining the eigenvectors of the combined Laplacians will be used to produce the optimal VICReg representation as formalized below.

\begin{theorem}
\label{thm:VICReg_optimal}
A global minimizer of the VICReg loss ($\alpha=\beta,\forall \gamma$) denoted by  $\mZ^*_{\alpha,\gamma}$ is obtained from \cref{eq:VICReg_optimal} along with the minimal achievable loss which are given by
\begin{align*}
    \mZ^*_{\alpha,\gamma}&=(\mP_{\alpha,\gamma}(\diag(\vlambda_{\alpha,\gamma}) N)^{1/2})_{:,1:K}\;\;\text{ and }\;\;
    \min_{\mZ \in \mathbb{R}^{N \times K}} \mathcal{L}_{\rm VIC}=\alpha (K-\|(\vlambda_{\alpha,\beta})_{1:K}\|_2^2),
\end{align*}
and any $K$-out-of-$N$ columns of $\mP_{\alpha,\gamma}(\mLambda_{\alpha,\gamma} N)^{1/2}$ is a local minimum of $\mathcal{L}_{\rm VIC}$. (Proof in \cref{proof:VICReg_optimal}.)
\end{theorem}

\begin{figure}[t!]
    \centering
    \begin{minipage}{0.01\linewidth}
    \rotatebox{90}{\scriptsize\color{blue}{$\min \mathcal{L}_{\rm VIC}$}\;\;\color{black}{/}\;\;\color{orange}{train acc. (\%)}}
    \end{minipage}
    \begin{minipage}{0.47\linewidth}
        \centering
        \begin{minipage}{0.49\linewidth}
        \centering
        \scriptsize
        $\alpha=1,\gamma=0$
        \end{minipage}
        \begin{minipage}{0.49\linewidth}
        \centering
        \scriptsize
        $\alpha=1,\gamma=0.1$
        \end{minipage}\\
        \includegraphics[width=\linewidth]{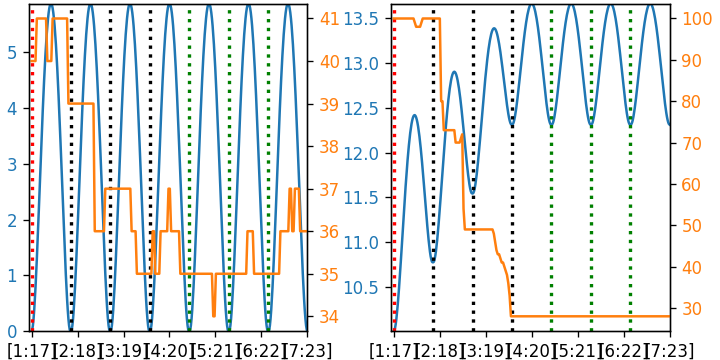}\\[-0.3em]
        {\scriptsize convex interp. between top $[k:K+k]$ eigenvectors of \cref{eq:VICReg_optimal}}
    \end{minipage}
    \begin{minipage}{0.02\linewidth}
    \rotatebox{90}{\scriptsize $\rank(\mG)=K$\hspace{.3cm}$\rank(\mG)=4$}
    \end{minipage}
    \begin{minipage}{0.235\linewidth}
    \centering
    \begin{minipage}{0.48\linewidth}
    \centering\scriptsize
    $\alpha=1,\gamma=0$
    \end{minipage}
    \begin{minipage}{0.48\linewidth}
    \centering\scriptsize
    $\alpha=1,\gamma=0.1$
    \end{minipage}\\
    \includegraphics[width=\linewidth]{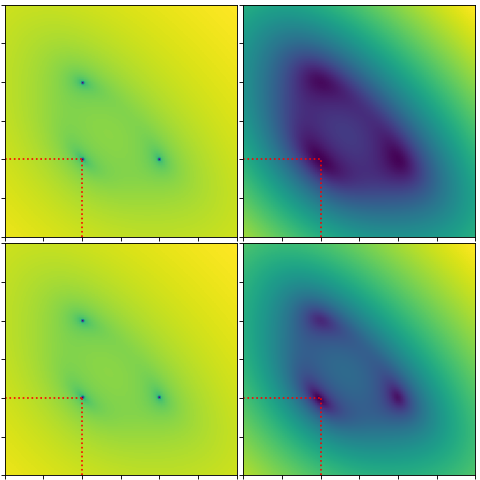}
    \end{minipage}
    \begin{minipage}{0.235\linewidth}
    \centering
    \begin{minipage}{0.48\linewidth}
    \centering\scriptsize
    $\alpha=1,\gamma=0$
    \end{minipage}
    \begin{minipage}{0.48\linewidth}
    \centering\scriptsize
    $\alpha=1,\gamma=0.1$
    \end{minipage}\\
    \includegraphics[width=\linewidth]{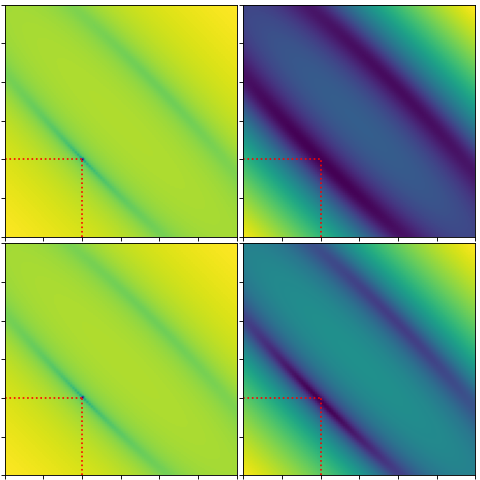}
    \end{minipage}
    \caption{\small {\bf Left:} depiction of the optimal VICReg loss with varying hyper-parameters ({\bf blue line}) when the representation is formed from the top $[k:K+k]$ eigenvectors of \cref{eq:VICReg_optimal} with convex interpolation in-between. Recall from \cref{thm:VICReg_optimal} that the global optimum is given by the $[1:K-1]$ case. We also depict the downstream task performance ({\bf orange line}) and we clearly observe that both are closely related as expected (see \cref{thm:optimal}). Notice that since we are considering classification, even without the correct first eigenvector the linear classifier on top of $\mZ^*_{\alpha,\gamma}$ is able to solve the task at hand thanks to the probability constraint that must sum to $1$ i.e. the last component can be recovers from the first $C-1$. {\bf Right:} depiction of the loss landscape of $\mathcal{L}_{\rm VIC}$ around the optimal $Z^*_{\alpha,\gamma}$ on the left using the directions provides by the top $[2:K+1]$ and $[3:K+2]$ eigenvectors of \cref{eq:VICReg_optimal}, and then with random directions in $\mZ$-space. All experiments employed $N = 256,K = 16,\rank(\mG)=4$.}
    \label{fig:VICReg_landscape}
\end{figure}

\begin{figure}[t!]
    \begin{minipage}{0.015\linewidth}
    \rotatebox{90}{\scriptsize\hspace{1.9cm}$k$\hspace{1.3cm} \color{blue}{$\min\mathcal{L}_{\rm vic}$} \color{black}{/}\color{orange}{train acc. (\%)}}
    \end{minipage}
    \begin{minipage}{0.67\linewidth}
    \centering
    \includegraphics[width=\linewidth]{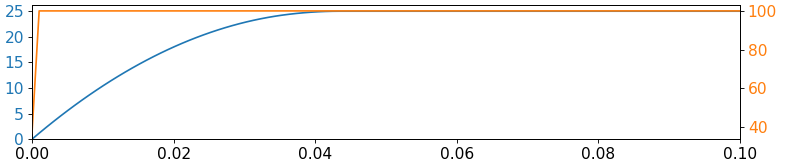}
    \includegraphics[width=\linewidth]{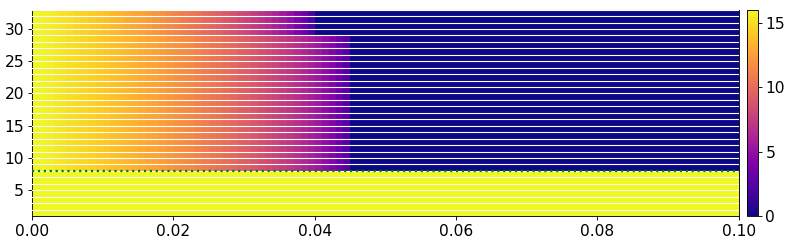}\\
    {\small invariance coefficient $\gamma$ ($\alpha=\beta=1$)}
    \end{minipage}
    \begin{minipage}{0.305\linewidth}
    \caption{\small {\bf Top:} the minimal VICReg loss ({\bf blue line}) and corresponding downstream task performance ({\bf orange line}). {\bf Bottom:} evolution of $\mZ^*_{\alpha,\gamma}$'s singular values. {\em We observe that VICReg benefits from a $(\alpha,\gamma)$-zone for which $\mZ^*_{\alpha,\gamma}$ remains full rank and still incorporates enough information on $\mG$ to solve the downstream task}. Hence, depending on the confidence one has into $\mG$, one can adjust $\gamma/\alpha$ to either incorporate no information on $\mG$, incorporate $\mG$ and maintain full-rank or entirely collapse to $\mG$. All experiments employed $N = 256,K = 32,\rank(\mG)=8$.}
    \label{fig:VICReg_landscape2}
    \end{minipage}
\end{figure}

\begin{figure}[t!]
    \centering
    \begin{minipage}{0.02\linewidth}
    \rotatebox{90}{\scriptsize $(\mathcal{L}_{\rm vic}(\mZ^{(t)})-\mathcal{L}_{\rm vic}(\mZ^*))^2$}
    \end{minipage}
    \begin{minipage}{0.45\linewidth}
    \centering
    {\small \hspace{0.5cm}K=8\hspace{1.5cm}K=32\hspace{1.5cm}K=64}\\
    \includegraphics[width=\linewidth]{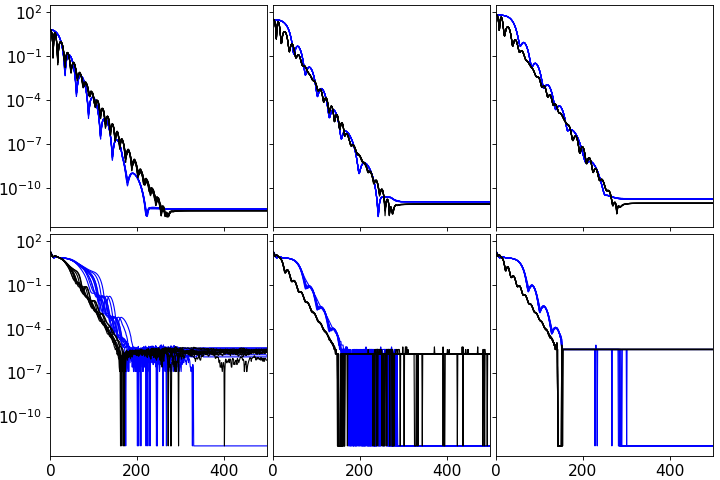}\\
    {\small \# iterations $(t)$}
    \end{minipage}
    \begin{minipage}{0.02\linewidth}
    \rotatebox{90}{\scriptsize $(\mathcal{L}_{\rm vic}(\mW^{(t)})-\mathcal{L}_{\rm vic}(\mW^*))^2$}
    \end{minipage}
    \begin{minipage}{0.45\linewidth}
    \centering
    {\small \hspace{0.5cm}K=8\hspace{1.5cm}K=32\hspace{1.5cm}K=64}\\
    \includegraphics[width=\linewidth]{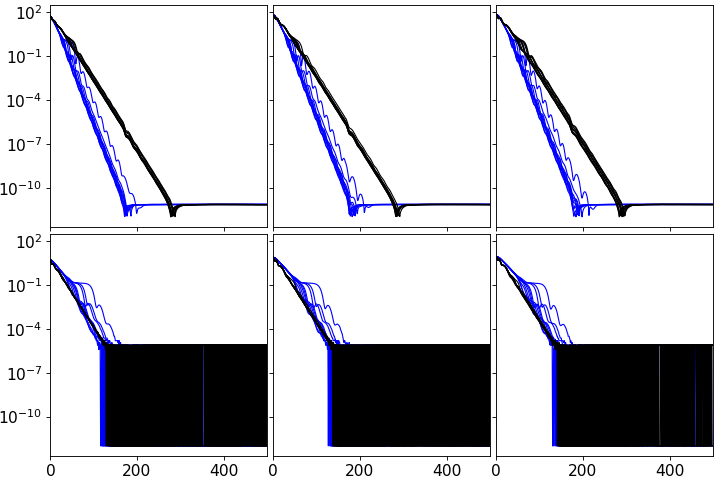}\\
    {\small \# iterations $(t)$}
    \end{minipage}
    \vspace{-0.3cm}
    \caption{\small {\bf Left:} depiction of the convergence between a randomly initialized representation $\mZ$ and the optimal one $\mZ^*$ from \cref{thm:VICReg_optimal} measured as quadratic difference of the VICReg loss $\mathcal{L}_{\rm vic}$. {\bf Right:} depiction of the convergence between a random initialized linear weight matrix $\mW$ and the optimal one $\mW^*$ from \cref{eq:Wstar_vicreg} also measured as quadratic difference of the VICReg loss $\mathcal{L}_{\rm vic}$. We employ $\alpha=1,\gamma=0$ on the {\bf top row} and $\alpha=1,\gamma=0.01$ on the {\bf bottom row} and N=512, C=8, D=64. In both cases, the {\bf blue line} corresponds to RMSProp optimizer and the {\bf black line} to SGD. Each setting is repeated $10$ times and each line corresponds to a different random initialization. We observe that in both settings gradient based learning is able to produce a final representation (left) and linear weight (right) that match with the optimum ones found analytically for VICReg.}
    \label{fig:convergence}
\end{figure}

The above result provides a few key insights. First, only the ratio $\gamma/\alpha$ governs the VICReg representation. Second, there exists many local minimum, some of which can be explicitly found by taking various $K$-out-of-$N$ columns of $\mP_{\alpha,\gamma}(\mLambda_{\alpha,\gamma} N)^{1/2}$ which we display in \cref{fig:VICReg_landscape} along with the loss landscape of $\mathcal{L}_{\rm VIC}$ around the optimal representation $\mZ^*$.
We also depict in \cref{fig:VICReg_landscape2} the evolution of the eigenvalues $(\vlambda_{\alpha,\gamma})_{1:K}$ for varying $\gamma$ along with the downstream task (induced by $\mG$) training performance. We observe that VICReg benefits from a sweet-spot where it can both preserve a full-rank representation $\mZ^*$ and incorporate enough information about $\mG$ to solve the task at hand perfect. As will become clear in \cref{sec:simclr,sec:BT} this does not hold for all methods as SimCLR and BarlowTwins collapse the rank of $\mZ^*$. We also depict on the left of \cref{fig:convergence} the convergence of the optimal representation of VICReg to the true one with different flavors of gradient descent.
Beyond interpretability, \cref{thm:VICReg_optimal} 
also clearly demonstrate the impact of $\mG$ on the quality of the obtained representation $\mZ^*_{\alpha,\gamma}$. We will make the ability of $\mZ^*_{\alpha,\gamma}$ to solve a downstream task at hand as a function of the spectrum of $\mG$ in \cref{sec:same_coin}. 

{\bf Computationally friendly solution.}~Before concluding this section we ought to recall that \cref{eq:VICReg_optimal}, which needs to be decomposed to obtain $\mZ^*_{\alpha,\gamma}$ is a dense $N \times N$ matrix e.g. as follows
\begin{lstlisting}[language=Python,escapechar=\%]
L = ch.diag(G.sum(1)) - G # Laplacian matrix %\lsComment{$\mL$}% in Eq. %\lsComment{\ref{eq:VICReg_optimal}}%
Lam, P = ch.linalg.eigh(torch.eye(N) - 1 / N - (gamma / alpha) * L) #equals to %\lsComment{(\ref{eq:VICReg_optimal})}%
Z_star = P[:, -K:] * ((Lam[-K:] * N).sqrt()) # equals to %\lsComment{$\mZ^*$ of Thm. \ref{thm:VICReg_optimal}}%
min_L = alpha * (K - Lam[-K:].norm().square()) # equals to %\lsComment{$\min \mathcal{L}_{\rm vic}$ of Thm. \ref{thm:VICReg_optimal}}%
\end{lstlisting}
However, we can equivalently find the optimal VICReg solution $\mZ^*_{\alpha,\gamma}$ only as a function of $\mL$ which is sparse e.g. for SSL $\mL$ contains $2N$ nonzero entries, and for supervised settings with $C$ balanced classes, $\mL$ contains $N^2/C$ nonzero entries. To do so, notice in \cref{eq:VICReg_optimal} that the left term $\mI-\frac{1}{N}\mathbf{1}\mathbf{1}^T$ mostly acts to ensure that the found eigenvectors with nonzero eigenvalues have $0$ mean. Instead, we can achieve the same goal by defining the following {\em sparse and symmetric} matrix
\begin{align}
    \begin{pmatrix}
    0 & \mathbf{1}^T\\
    \mathbf{1} & \mI - \frac{\gamma}{\alpha}\mL
    \end{pmatrix}\in\mathbb{R}^{ (N+1)\times(N+1)},\label{eq:sparse_J}
\end{align}
and by taking the top $[2:K]$ (and not $[1:K-1]$ as with \cref{eq:VICReg_optimal}) eigenvectors. In short, while \cref{eq:VICReg_optimal} was adding a connection from each node of the graph to all others, \cref{eq:sparse_J} introduces a new node to the graphs and connects it to all others, akin to the Hubbard-Stratonovich transformation \citep{hirsch1983discrete}. The great advantage of \cref{eq:sparse_J} is the sparsity of the matrix allowing it to be easily stored and for which there exists efficient routines to rapidly obtain the top eigenvectors, even for large $N$. One popular solution is the Locally Optimal Block Preconditioned Conjugate Gradient \citep{knyazev1987convergence} which only needs to evaluate matrix-vector products.

Prior to moving to other SSL methods, we first emphasize the ability of VICReg to recover {\em local} spectral methods in the following sections.

\subsection[VICReg Recovers Laplacian Eigenmaps]{VICReg Recovers Laplacian Eigenmaps in Feature Space and Kernel Locality Preserving Projection in Data Space}
\label{sec:vicreg_le}

From the previous section \cref{thm:VICReg_optimal,eq:VICReg_optimal} we obtained that the ratio $\frac{\gamma}{\alpha}$ entirely controls the mixing between the Laplacian matrix of a complete graph and the one from $\mG$. In other words, this ratio controls the spectral properties of $\mZ^*_{\alpha,\gamma}$ by interpolating between the ones of $\mI\frac{1}{N}\mathbf{1}\mathbf{1}^T$ and the ones of $\mL$ (recall \cref{fig:VICReg_landscape2}). The goal of this section is to study more precisely the scenario where $\alpha \gg \gamma$ i.e. we prioritize the variance-covariance terms, in which case VICReg recovers Laplacian Eigenmaps (LE) \citep{belkin2003laplacian} in feature space, and kernel Locality Preserving Projection (kLPP) \citep{he2003locality} in data space.

{\bf In feature space.}~LE is a non-parametric method searching for a representation $\mZ$ by minimizing the following Brockett \citep{brockett1991dynamical} optimization problem
\begin{align}
    \min_{\mZ}\Tr\left(\mZ^T \left(\mD-\mG\right)\mZ\right) \text{ s.t. }\mZ^T\mD\mZ=\mI,\label{eq:LE}
\end{align}
with $\mD$ the diagonal degree matrix of $\mG$ (recall \cref{eq:VICReg_optimal}).
The relation between \cref{eq:LE} and VICReg comes by combining the following two facts. First, minimizing \cref{eq:LE} is done by taking the eigenvectors associated to the $K$ smallest eigenvalues of $\mI-\mD^{-1}\mG$ (see \citet{liang2021generalizing} and Corollary 4.3.39 from \citet{horn2012matrix}) yet, ---LE disregards the eigenvector associated to the eigenvalue $0$--- to produce the optimal LE solution denoted as $\mZ^*_{\rm LE}$. Doing so, \citet{belkin2003laplacian} produced a representation whose columns have $0$-mean since the only eigenvector with nonzero mean is the one associated to the eigenvalue $0$ (details in \cref{proof:vicreg_le}). Second, VICReg employs a graph $\mG$ for which $\mD=c\mI$ with $c=2$ (positive pairs), $c=3$ (positive tirplets, and so on. And thus, the eigenvectors associated to the nonzero eigenvalues of $\mD-\mG$ or $\mI-\mD^{-1}\mG$ are the same i.e. the optimal solution of the LE problem minimizes the VICReg criterion with strict enforcement that the variance+covariance terms, denoted as $\mathcal{L}_{var},\mathcal{L}_{\rm cov}$ hereon, are zero.

\begin{theorem}
\label{thm:LE}
Given a dataset $\mX$ and relation matrix $\mG$, minimizing the VICReg loss (\cref{eq:VICReg}) with constraint that the variance and covariance loss are $0$ ($\alpha,\beta$ become irrelevant) as in
\begin{align*}
    \mathcal{L}_{\rm vic}(\mZ^*_{\rm LE})=\min_{\mZ \in \mathbb{R}^{N \times K}} \mathcal{L}_{\rm inv}(\mZ) \text{ s.t. } \mathcal{L}_{\rm var}=0 \text{ and }\mathcal{L}_{\rm cov}=0.
\end{align*}
(Proof in \cref{proof:vicreg_le}.)
\end{theorem}

Hence, given a relation matrix $\mG$, solving the LE problem is equivalent to solving a constrained VICReg problem. We ought to highlight however that a crucial part of LE lies in the design of that matrix $\mG$, often found from a $k$-NN graph \citep{hautamaki2004outlier} of the samples $\mX$ in the input space, while in SSL it is constructed from data-augmentations or given.

Going beyond LE, one can easily obtain that if $\left(\mD-\mG\right)$ is idempotent ---which is true for common supervised and SSL scenarios as $\mG$ corresponds to a union of complete graphs--- then LE also corresponds to Locally Linear Embedding \citep{roweis2000nonlinear} and so does VICReg.

{\bf In data space.}~One difficulty arising from LE, and from non-parametric methods in general, is the ability to produce new representations $\vz$ for new data samples that were not present when solving \cref{eq:LE}. This led to the development of a two-step modeling process as 
\begin{align*}
    \vx \; (\in \mathbb{R}^{D}) \mapsto \vh=\phi(\vx)\; (\in \mathbb{R}^{S}) \mapsto \vz=\mW^T\vh (\in \mathbb{R}^{K}),
\end{align*}
with $S \gg K, \mW \in \mathbb{R}^{S \times K}$, and where the first mapping 's goal is to learn a generic input embedding that can be reused on new samples $\vx$. To see this, let's collect those mappings for all the training set into the $N\times S$ matrix $\Phi$. With that, the LE problem in data space ---known as the Kernel Locality Preserving Projection problem--- becomes
\begin{align}
    \min_{\theta:\mW^T\Phi^T\mD\Phi\mW=\mI}\Tr\left(\mW^T \Phi^T\left(\mD-\mG\right)\Phi\mW\right),\label{eq:KLPP}
\end{align}
so that the original LE representation $\mZ$ can be obtained simply as $\mZ=\Phi\mW$. And more importantly, given a new sample $\vx$, one can simply obtain the new representation via $\vz=\mW^T\phi(\vx)$.

Most existing solutions have been leveraging the Moore-Aronszajn theorem \citep{aronszajn1950theory} and the fact that we are in a finite data regime to shift the question {\em what nonlinear and high-dimensional operator $\phi$ should be used?} to the perhaps simpler question {\em what symmetric positive (semi-)definite matrix $\Phi \Phi^T$, i.e. what kernel, function should be used?} \citep{bengio2003out,cheng2005supervised,tai2022kernelized}. Doing so, $\mW^T\phi(\vx)$ can now be easily implemented e.g. as a Generalized Regression Network \citep{broomhead1988radial,specht1991general}. The crucial result of interest for our study is the following one that combine \cref{thm:LE} with a result from \citet{he2003locality} demonstrating the equivalence between LE in feature space and KLLE in data space.

\begin{proposition}
VICReg with variance/covariance constraint solves LE in embedding space and KLLE in input space (recall \cref{eq:KLPP}) employing a DN for $\mW^T \circ \phi$.
\end{proposition}

Equipped with those results, we can now propose a practical result by obtaining in closed-form the optimal parameter weights that solve the constrained VICReg criterion with a linear network.

\subsection[VICReg Recovers Locality Preserving Projections and Linear Discriminant Analysis]{With a Linear Network VICReg Recovers Locality Preserving Projections and Linear Discriminant Analysis with Analytical Optimal Network Parameters}
\label{sec:vicreg_linear}

The goal of this section is to focus on the linear case i.e. $\mZ=\mX\mW$. In that setting, and still employing the squared variance term as in \cref{thm:vicreg_linear}, VICReg recovers two known spectral methods: Locality Preserving Projections (LPP) \citep{he2003locality} for an arbitrary relation matrix $\mG$, and Linear Discriminant Analysis (LDA) \citep{fisher1936use,cohen2014applied} when $\mG$ is the supervised relation matrix. In both cases we obtain the analytical form of the optimal weights $\mW$.

Prior connecting linear VICReg to LPP and LDA we ought to highlight the optimal weights $\mW^*$ of the linear VICReg model (derivations in \cref{proof:linear_vicreg}) given by
\begin{align}
    \mW^* = \left(\Cov(\mX)^{-1}\left(\Cov(\mX)-\frac{\gamma}{\alpha N}\mX^T\mL\mX\right)^{\frac{1}{2}}\right)_{.,1:K},\label{eq:Wstar_vicreg}
\end{align}
which is easily computed as
\begin{lstlisting}[language=Python,escapechar=\%]
cov = torch.cov(X.t(), correction=0) # cov. w/o correction (otherwise scale L by (N-1)
B = cov - (gamma/ (N * alpha)) * (X.t() @ L @ X)
Wstar = torch.linalg.solve(A, sqrtm(B)) # numerically stable solution of %\lsComment{\ref{eq:Wstar_vicreg}}%, sqrtm(B)=%\lsComment{\scriptsize$\sqrt{B}$}%
\end{lstlisting}
and we depict the gradient convergence towards this optimal solution is depicted on the right of \cref{fig:convergence}.

We now turn to connecting linear VICReg to LPP and LDA, heavily relying on \cref{thm:LE}. In fact, we already saw that VICReg with a DN corresponds to the Kernel LE methods. In the linear regime, it is already established that kernel LE recovers LPP. The perhaps less direct result concerns the recovery of LDA for which we first need to introduce the supervised counterpart of $\mG$. In that case, $\mG$ describes the known class relation from $\vy \in \{1,\dots,C\}^N$ as in
\begin{align}
    (\mG_{\rm s})_{i,j}=1_{\{(\vy)_i=(\vy)_j\}} 1_{\{i \not = j\}},\;\;\;\text{(supervised $\mG$)}\label{eq:H}
\end{align}
which can be directly computed as
\begin{lstlisting}[language=Python,escapechar=\%]
Gs = torch.eq(target, target[:, None]).float() - torch.eye(N) # equals to Eq. %\lsComment{\ref{eq:H}}%
\end{lstlisting}
Note that the degree matrix of $\mG_{\rm s}$ will contain for each diagonal entry $i$ the number of samples that belong to the class $(\vy)_{i}$ of sample $i$. 

\begin{theorem}
\label{thm:vicreg_linear}
Variance-covariance constrained VICReg (recall \cref{thm:LE}) with a linear network ($K\leq D$) recovers LPP, and LDA (with $\mG_{\rm s}$ from \cref{eq:H} and $K=C$). In both cases the optimal parameter $\mW^*$ is given by the top-$K$ eigenvectors of $(\mX^T(\mD-\mG)\mX)^{-1}\mX^T\mG\mX$ with $\mG \mapsto \mG_{\rm s}$ for LDA.
(Proofs in \cref{proof:vicreg_lpp,proof:vicreg_lda}.)
\end{theorem}

 Interestingly, the eigenvalues associated to the eigenvectors of $\mW^*$ exactly recover the multivariate analysis of variance (MANOVA) sufficient statistics of the data \citep{o1985manova,huberty2006applied}. Hence, although not further explored in this study, we believe that important statistical results could be further obtained e.g. to assess the goodness-of-fit of the model without requiring a downstream task \citep{mika1999fisher,ioffe2006probabilistic,xue2014does}.

After a thorough analysis of VICReg we now propose to turn to another important SSL loss which is SimCLR, and its variants.

\section[SimCLR]{SimCLR/NNCLR/MeanShift Solve a Generalized Multidimensional Scaling Problem \`a la ISOMAP}
\label{sec:simclr}

Recall from \cref{sec:background,eq:simCLR} that the SimCLR loss first computes a similarity matrix $\widehat{\mG}$ of some flavor that depends on the representation $\mZ$, and then compares it against the known data relation $\mG$. Different $\mG$ matrices lead to different variants of SimCLR \citep{chen2020simple} such as NNCLR \citep{dwibedi2021little} or MeanShift \citep{koohpayegani2021mean}, hence making our results general.
The goal of this section is two-fold. First, we demonstrate that different $\mZ \mapsto \widehat{\mG}$ mappings solve different optimization problems (\cref{lemma:infoNCE}) ---all trying to estimate the similarity matrix $\mG$ from the signals $\mZ$ akin to Laplacian estimation in Graph Signal Processing. Second, we demonstrate that SimCLR and its variants force $\mZ$'s spectrum to align with the one of $\mG$ (through $\widehat{\mG}$) akin to the ISOMAP method, by means of a Multi-Dimensional Scaling solution (\cref{thm:simclr_mds}).

\begin{figure}[t!]
    \centering
    \begin{minipage}{0.35\linewidth}
    \includegraphics[width=\linewidth]{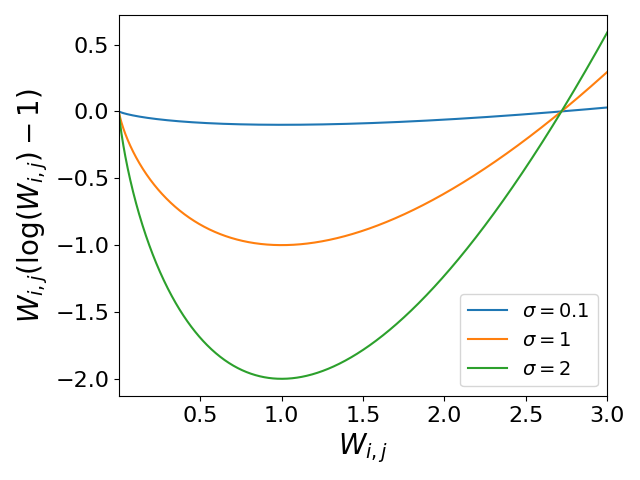}
    \end{minipage}
    \begin{minipage}{0.35\linewidth}
    \includegraphics[width=\linewidth]{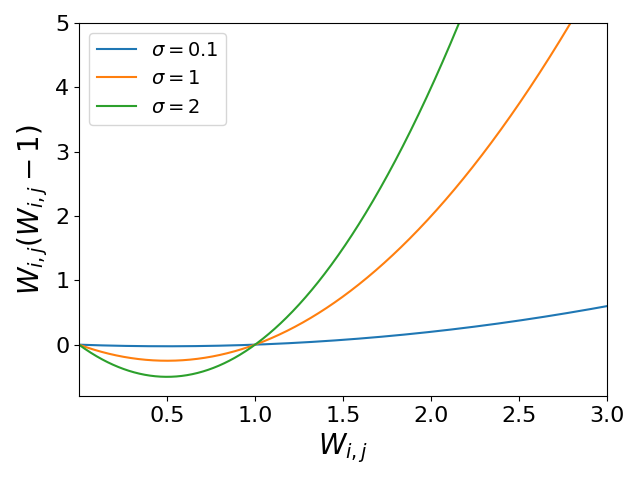}
    \end{minipage}
    \begin{minipage}{0.28\linewidth}
    \caption{\small Depiction of the regularization term $\mathcal{R}_{\rm log}$ ({\bf left}) and $\mathcal{R}_{\rm log}$ ({\bf right}) from \cref{eq:R_C} for varying value of $(\mW)_{i,j}$ and three different temperatures $\tau \in \{0.1,1,2\}$ demonstrating how such regularization prevents $\widehat{\mG}$ to collapse to the trivial $0$ matrix in the optimization problem in \cref{eq:opti_W}.}
    \label{fig:regularization}
    \end{minipage}
\end{figure}

\subsection[Step 1]{Step 1: SimCLR Pairwise Similarities Solve a Graph Laplacian Estimation Problem}
\label{sec:simclr_first}

Let's first define the minimization problem that given a set of signals i.e. rows of $\mZ$ produces a relation estimate $\widehat{\mG}$ of $\mG$. To ease notations, we gather in the $N \times N$ matrix $\mD$ all the pairwise distances $(\mD)_{i,j}=d(f_{\theta}(\vx_i),f_{\theta}(\vx_j))$ with $d$ any preferred metric. The standard problem of estimating $\widehat{\mG}$ from $\mZ$ can be cast as an optimization problem \citep{dong2016learning,kalofolias2016learn} as
\begin{align}
    \widehat{\mG}_{d,\mathcal{R}}&=\argmin_{\mG \in \mathcal{G}}\sum_{i,j}d(f_{\theta}(\vx_i),f_{\theta}(\vx_j))(\mG)_{i,j}+\mathcal{R}(\mG)
    =\argmin_{\mG \in \mathcal{G}}\Tr(\mD\mG)+\mathcal{R}(\mG),\label{eq:opti_W}
\end{align}
with $\mathcal{G}$ the set (or subset) of symmetric matrices with nonnegative entries and zero diagonal, and with $\mathcal{R}$ a regularizer preventing $\widehat{\mG}$ to be the trivial zero matrix e.g.
\begin{align}
    \mathcal{R}_{\rm log}(\mG)=\sum_{i\not =j}\tau \mG_{i,j}(\log(\mG_{i,j})-1)\;\;\text{ or }\;\;
    \mathcal{R}_{\rm F}(\mG)=\sum_{i\not =j}\tau \mG_{i,j}(\mG_{i,j}-1).\label{eq:R_C}
\end{align}
We provide in \cref{fig:regularization} a depiction of the impact of $\mathcal{R}(\mG)$ which pushes the entries of the weight matrix to be close to $1$ with strength depending on the temperature parameter $\tau$.

Hence $\widehat{\mG}$ from \cref{eq:opti_W} is the optimal graph ---expressed as a weight matrix--- for which the signal $\mZ=f_{\theta}(\mX)$ on that graph is smooth. For example on can solve \cref{eq:opti_W} only on $\mathcal{G}_{\rm rsto}$, the space of right-stochastic matrices i.e. a subset of $\mathcal{G}$ that only contains matrices whose rows sum to $1$ i.e. $\mathcal{G}_{\rm rsto}=\{\mG \in \mathcal{G}:\mG \mathbf{1}=\mathbf{1}\}$.
We now propose the first formal result that ties the estimate $\widehat{\mG}$ from \cref{eq:opti_W} to the one of SimCLR/NNCLR/MeanShift whenever the regularizer $\mathcal{R}$ is given by
but with different spaces for $\mG$.

\begin{theorem}
\label{lemma:infoNCE}
Using $\mathcal{R}$ from \cref{eq:R_C} leads to the following graph weight estimate
\begin{align}
    (\widehat{\mG}_{d,\mathcal{R}_{\rm log}})_{i,j}&=e^{\frac{-1}{\tau}d(f_{\theta}(\vx_i),f_{\theta}(\vx_j))}1_{\{1 \not = j\}},&&\text{(with $\mathcal{G}$)}\nonumber\\
    (\widehat{\mG}_{d,\mathcal{R}_{\rm log}})_{i,j}&=\frac{e^{\frac{-1}{\tau}d(f_{\theta}(\vx_i),f_{\theta}(\vx_j))}}{\sum_{j\not = i} e^{\frac{-1}{\tau}d(f_{\theta}(\vx_i),f_{\theta}(\vx_j))}}1_{\{1 \not = j\}},&&\text{(with $\mathcal{G}_{\rm rsto}$)}\label{eq:opti_W_simclr}
\end{align}
and thus if $d$ is the cosine distance, \cref{eq:opti_W_simclr} recovers SimCLR similarity estimate, and if $d$ is the $\ell_2$ distance, \cref{eq:opti_W_simclr} recovers the Lifted Structured Loss \citep{weinberger2009distance}. (Proof in \cref{proof:contrastive}.)
\end{theorem}

\begin{figure}[t!]
    \centering
    \begin{minipage}{0.65\linewidth}
        \centering
        \hfill
        \begin{minipage}{0.018\linewidth}
            \rotatebox{90}{\small$(\mSigma_Z)_{k,k}$ in log-scale}
        \end{minipage}
        \begin{minipage}{0.972\linewidth}
            \centering
            \hfill
            \begin{minipage}{0.23\linewidth}\small
            \centering
            cosine+$\mathcal{R}_{\rm F}$
            \end{minipage}
            \begin{minipage}{0.23\linewidth}\small
            \centering
            cosine+$\mathcal{R}_{\rm log}$\\(SimCLR/NNCLR)
            \end{minipage}
            \begin{minipage}{0.23\linewidth}\small
            \centering
            $\ell_2+\mathcal{R}_{\rm F}$
            \end{minipage}
            \begin{minipage}{0.23\linewidth}\small
            \centering
            $\ell_2+\mathcal{R}_{\rm log}$
            \end{minipage}\\
            \includegraphics[width=\linewidth]{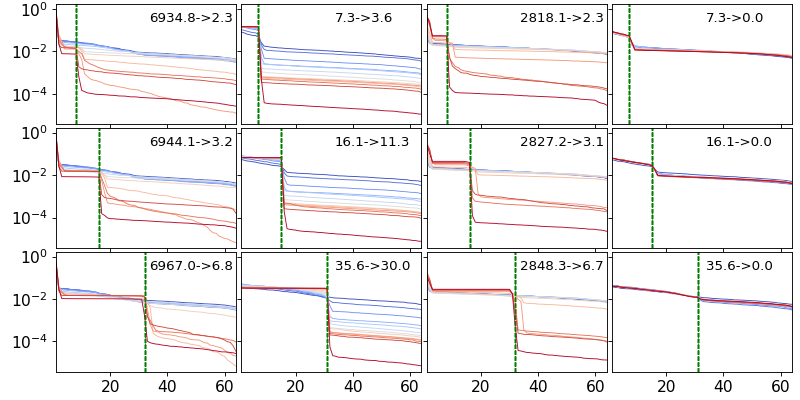}\\[-0.5em]
            \hfill
            \begin{minipage}{0.225\linewidth}\small
            \centering
            $k$
            \end{minipage}
            \begin{minipage}{0.225\linewidth}\small
            \centering
            $k$
            \end{minipage}
            \begin{minipage}{0.225\linewidth}\small
            \centering
            $k$
            \end{minipage}
            \begin{minipage}{0.225\linewidth}\small
            \centering
            $k$
            \end{minipage}
        \end{minipage}
    \end{minipage}
    \begin{minipage}{0.34\linewidth}
    \caption{\small Depiction of the singular values $\mSigma_{Z}$ of the representation $\mZ$ learned by the SimCLR/NNCLR loss with $\mathcal{G}_{\rm rsto}$, varying $\rank(\mG) \in \{8,16,32\}$ ({\bf rows, green dotted lines}) and evolution during training ({\bf from blue to red}, number in top-right corner) with various ($d,\mathcal{R}$) configurations ({\bf columns}, recall \cref{lemma:infoNCE}). We observe that the rank of the learned representation matches exactly the one of $\mG$  validating the result from \cref{thm:simclr} regardless of the chosen hyper-parameters, making SimCLR's performances more sensitive to the design of $\mG$ than methods like VICReg whose representation preserves remains full-rank (recall \cref{fig:VICReg_landscape2}).}
    \label{fig:simclr}
    \end{minipage}
\end{figure}

The above result plays a crucial role as it motivates the need to better understand SSL from a theoretical perspective.
Based on this maximization principle, we can now obtain novel variations of SimCLR by solving \cref{eq:opti_W} with different constraints. For example one could obtain (derivations in \cref{proof:contrastive}) the following variations that are akin to the quantities used in \citet{hadsell2006dimensionality,weinberger2009distance} i.e. with $d$ the $\ell_2$ distance and no exponentiation
\begin{align}
\widehat{\mG}_{d,\mathcal{R}_{\rm F}}&=\relu\left(\mathbf{1}\mathbf{1}^T-\mI-\mD/\tau\right),&&\text{(with $\mathcal{G},\tau >0$)},\label{eq:simclr_G1}\\
\widehat{\mG}_{d,\mathcal{R}_{\rm F}}&=\relu\left(\frac{1}{N-1}\left(\mathbf{1}\mathbf{1}^T-\mI\right)-\frac{1}{\tau}\mD\left(\mI - \frac{1}{N-1}\mathbf{1}\mathbf{1}^T\right)\right),&&\text{(with $\mathcal{G}_{\rm rsto},\tau >0$)}.\label{eq:simclr_G2}
\end{align}
Now that we have demonstrated how SimCLR estimates the graph at hand, the next section will focus on the second step: matching $\widehat{\mG}$ to $\mG$.



\subsection[Step 2]{Step 2: SimCLR Fits the Estimated Graph $\widehat{\mG}$ to the Known Graph $\mG$}
\label{sec:simclr_second}

SimCLR's first step produced a graph estimate $\widehat{\mG}$ from $\mZ$. The second step now consists in measuring how fit is this estimate compared to $\mG$, so that the model's parameters can be tuned to increase this fitness. We demonstrate in this section that in doing so, SimCLR forces $\mZ$ to have the same nonzero left singular vectors as the nonzero eigenvectors of $\mG$, and that as opposed to VICReg, the rank between $\mZ$ and $\mG$ matches.

The method used to compare $\widehat{\mG}$ and $\mG$ should reflect the properties fulfilled by those matrices e.g. being doubly-stochastic or right-stochastic and the type of measure one aims to impose. In all generality, let's denote the SimCLR comparison method to be one of the two following variants (depending on the type of constraints put on $\widehat{\mG}$ and $\mG$
\begin{align}
    \mathcal{L}_{\rm SimCLR}=\begin{cases}
    \| \mG - \widehat{\mG}(\mZ)\|_{F}^2, & \text{(Euclidean)}\\
    -\Tr\left(\left(\diag(\mG\mathbf{1})^{-1}\mG\right)^T\log\left(\diag(\widehat{\mG}(\mZ)\mathbf{1})^{-1}\widehat{\mG}(\mZ)\right)\right), & \text{(Cross-Entropy)}
    \end{cases},\label{eq:simCLR2}
\end{align}
where the $\log$ is applied element-wise. When minimizing \cref{eq:simCLR2}, SimCLR learns an embedding $\mZ$ so that its graph estimate $\widehat{\mG}$ matches closely the known graph $\mG$. To formalize below the form of the optimal SimCLR representation, we first ought to recall that we denote by $\mU_{G}$ and $\mSigma_{G}$ the left singular vectors and singular values of $\mG$ respectively. Notice that since $\mG$ is symmetric semi-definite positive, $\mU$ also corresponds to its eigenvectors and $\mSigma_{G}^2$ to its eigenvalues.

\begin{theorem}
\label{thm:simclr}
A global minimizer of the SimCLR loss denoted by $\mZ^*$ along with the minimal achievable loss using \cref{eq:simclr_G1} or \cref{eq:simclr_G2}, $\tau \geq \max_{i,j}(D)_{i,j}$ and with the LHS of \cref{eq:simCLR2} are given by
\begin{align*}
    \mZ^*_{\tau}&=(\mU_{G}\mSigma_{G}^{1/2})_{:,1:K}\;\;\text{ and }\;\;
    \min_{\mZ \in \mathbb{R}^{N \times K}} \mathcal{L}_{\rm SimCLR}=\sum_{k=K+1}^{N}(\mSigma_{G}^2)_{k,k},
\end{align*}
up to permutations of the singular vectors associated to the same singular value, and regardless of the loss and graph estimation, the rank of $\mZ^*_{\tau}$ is given by $\min(K,\rank(\mG))$. (Proof in \cref{proof:simclr}.)
\end{theorem}

We illustrate the above theorem for many combinations of distances and regularizers in \cref{fig:simclr} where we see that in all cases, SimCLR forces the representations $\mZ$ to have a dimensional collapse, a phenomenon first observed in \citet{hua2021feature} and that has been one of the unanswered phenomenon in SSL \citep{arora2019theoretical,tosh2021contrastive}. In our goal to unify SSL methods under the realm of spectral embedding methods, we now propose the following section that ties SimCLR and its variants to {\em global} spectral methods.

\subsection{SimCLR is Akin to ISOMAP in Feature Space and Kernel ISOMAP in Input Space}
\label{sec:simclr_isomap}

We now propose to tie the SimCLR method along with its variants e.g. NNCLR to known {\em global} spectral methods, e.g. ISOMAP \citep{tenenbaum2000global} contrasting from VICReg which was tied to {\em local} spectral methods (recall \cref{sec:vicreg_le,sec:vicreg_linear}).

{\bf In feature space.}~Let's first recall that ISOMAP is a variation of Multi-Dimensional Scaling (MDS) \citep{kruskal1964multidimensional} also known as Principal Coordinates Analysis. Classical MDS tries to learn embedding vectors that have similar pairwise distance (usually $\ell_2$) than the pairwise distance of the given input data. Often, MDS does this by using similarities instead of distances and thus by solving the following optimization problem $\min_{\mZ}\|\mG - \mZ\mZ^T \|_F^2$,
where $\mG$ is the Gram matrix of the inputs i.e. $\mG=\mX\mX^T$. At the most general level, ISOMAP simply corresponds to solving that same optimization problem but after redefining $\mG$ to better capture the geometric information of $\mX$ e.g. using the shortest path distance of the $k$-NN graph of $\mX$ \citep{preparata2012computational}. The surprising result that we formalize below is that SimCLR and in particular NNCLR recover ISOMAP, the former by prescribing $\mG$ given the known positive pairs, and the latter by building a nearest neighbor graph.

\begin{proposition}
\label{thm:simclr_mds}
Using the settings of \cref{thm:simclr}, SimCLR recovers a Generalized MDS akin to ISOMAP (and MDS iff $\mG=\mX\mX^T$). (Proof in \cref{proof:simclr_mds}.)
\end{proposition}

{\bf In data space with DNs.}~From the above, we can extend \cref{thm:simclr_mds} but in input space, in a very similar way as was done in \cref{sec:vicreg_le}. In fact, originating in \citet{webb1995multidimensional}, there was a search to extend MDS, and ISOMAP to an input space formulation to solve the out-of-bag problem. In this setting, and taking MDS as an example, the original similarity matrix $\mZ \mZ^T$ is replaced with $\Phi \mW^T \mW \Phi^T$  using the same notations as in \cref{eq:KLPP} and already known relationship between those models, we obtain the following.

\begin{proposition}[\citep{williams2000connection}]
Whenever SimCLR recovers ISOMAP or MDS in feature space, it recovers kernel ISOMAP or kernel PCA \citep{scholkopf1998nonlinear} in input space.
\end{proposition}



We will now turn to BarlowTwins, another non-contrastive method akin to VICReg (both of which fall back to LDA in the linear regime and with supervised $\mG$).

\section[BarlowTwins]{BarlowTwins Solves a (Kernel) Canonical Correlation Analysis Problem and Can Recover VICReg}
\label{sec:BT}

Our last step in our journey to unify SSL methods under spectral embedding methods deals with BarlowTwins. Akin to the development for VICReg and SimCLR, BarlowTwins will also fall back to a known spectral method in embedding space (\cref{sec:BT_first}) and in data space (\cref{sec:barlowtwins_linear}) where in the later case we again obtain the close-form optimal network parameters in the linear regime.

\subsection[Stationary signal on G]{BarlowTwins Recovers Kernel Canonical Correlation Analysis}
\label{sec:BT_first}

Recall from \cref{sec:background,eq:BT} that the BarlowTwins loss is based on a cross-correlation matrix between positive pairs of samples. As we did for VICReg and SimCLR, our goal here is to tie BarlowTwins to a known spectral method known as Kernel Canonical Correlation Analysis.

There exists many different ways to formulate the CCA problem, we present one here in the linear regime to simplify notations based on \cite{cunningham2015linear}. The goal of (linear) CCA is to learn pairs of filters that produce maximally correlated features as in
\begin{align}
    \max_{\mW_{\rm a}\in\mathbb{R}^{D_{\rm a}\times K},\mW_{\rm b}\in\mathbb{R}^{D_{\rm b}\times K}}\frac{\Tr(\mW_{\rm a}^T\mSigma_{ab}\mW_{\rm b})}{\sqrt{\Tr(\mW_{\rm a}^T\mSigma_{aa}\mW_{\rm a})\Tr(\mW_{\rm b}^T\mSigma_{bb}\mW_{\rm b})}}, \text{ s.t. }\begin{cases}
    \frac{1}{N}\mW_{\rm a}^T\mSigma_{aa}\mW_{\rm a}=\mI\\
    \frac{1}{N}\mW_{\rm b}^T\mSigma_{bb}\mW_{\rm b}=\mI\\
    \mW_{\rm a}^T\mSigma_{ab}\mW_{\rm b}=\mLambda,
    \end{cases}\label{eq:CCA}
\end{align}
with $\mLambda$ a diagonal $(K \times K)$ matrix, and with $\mSigma_{\rm aa}=\mX_{\rm a}^T\mX_{\rm a},\mSigma_{\rm ab}=\mX_{\rm a}^T\mX_{\rm b}$ and so on. We thus observe that \cref{eq:CCA} is the BarlowTwins objective of \cref{eq:BT} up to rescaling of the weight matrices, as CCA aims to make $\mW_{\rm a}^T\mSigma_{ab}\mW_{\rm b}=\mW_{\rm a}^T\mX_{\rm a}^T\mX_{\rm b}\mW_{\rm b}=\mZ_{\rm a}^T\mZ_{\rm b}=\mLambda$ diagonal with maximal diagonal entries.

Going to the nonlinear regime i.e. kernel CCA follows directly be employing $\phi(\vx)$ embeddings of the input observations in-place of the inputs. We thus obtain the following result that nicely parallels with the ones we obtained for VICReg and SimCLR. In data space, BarlowTwins can be regarded (put in perspective with \cref{sec:barlowtwins_linear}) as a nonlinear canonical correlation analysis (NLCA) \citep{dauxois1998nonlinear} and in particular Kernel CCA (KCCA) \citep{lai2000kernel,fukumizu2007statistical} akin to how VICReg recovered Kernel Locality Preserving Projection and SimCLR Kernel ISOMAP. We leverage the same notations as in \cref{sec:vicreg_le}.

\begin{theorem}
\label{thm:BT}
BarlowTwins recovers Kernel Canonical Correlation Analysis with a DN as the featurizer $\phi$ and produce a representation with rank $\min(K,D,\rank(\mG))$. (Proof in \cref{proof:BT}.).
\end{theorem}

In addition to the above, we can obtain further interpretation of the components of BarlowTwins loss. For example, notice that \cref{eq:CCA} is only well-defined if $K<\min(D_{\rm a},D_{\rm b})$. This limitation led to ridge-type CCA regularization which as been introduced in \citet{gretton2005kernel} as a mean to introduce numerical stability which in the context of linear CCA has been introduced by \citet{vinod1976canonical} under the name {\em canonical ridge} and recovers exactly the addition of the $\epsilon$ constant in the denominator of the BarlowTwins loss in \cref{eq:BT}.

\begin{figure}[t!]
    \centering
    \begin{minipage}{0.65\linewidth}
        \centering
        \hfill
        \begin{minipage}{0.018\linewidth}
            \rotatebox{90}{\small$(\mSigma_Z)_{k,k}$ in log-scale}
        \end{minipage}
        \begin{minipage}{0.972\linewidth}
            \centering
            \hfill
            \begin{minipage}{0.23\linewidth}\small
            \centering
            \scriptsize
            $\rank(\mZ_{\rm init})=4$
            \end{minipage}
            \begin{minipage}{0.23\linewidth}\small
            \centering
            \scriptsize
            $\rank(\mZ_{\rm init})=8$
            \end{minipage}
            \begin{minipage}{0.23\linewidth}\small
            \centering
            \scriptsize
            $\rank(\mZ_{\rm init})=16$
            \end{minipage}
            \begin{minipage}{0.23\linewidth}\small
            \centering
            \scriptsize
            $\rank(\mZ_{\rm init})=64$
            \end{minipage}\\
            \includegraphics[width=\linewidth]{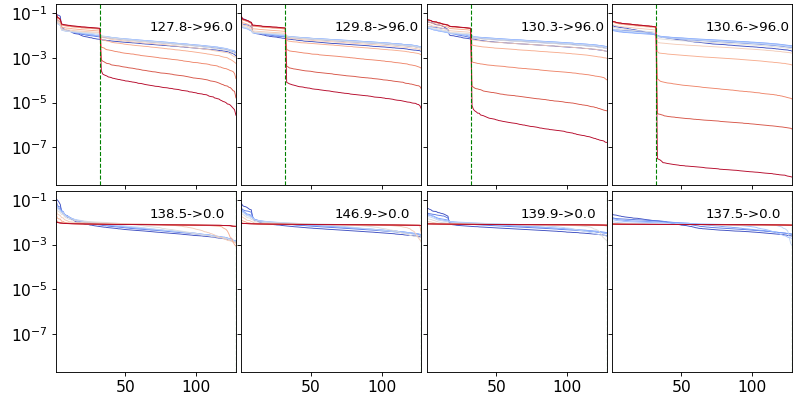}\\[-0.5em]
            \hfill
            \begin{minipage}{0.225\linewidth}\small
            \centering
            $k$
            \end{minipage}
            \begin{minipage}{0.225\linewidth}\small
            \centering
            $k$
            \end{minipage}
            \begin{minipage}{0.225\linewidth}\small
            \centering
            $k$
            \end{minipage}
            \begin{minipage}{0.225\linewidth}\small
            \centering
            $k$
            \end{minipage}
        \end{minipage}
    \end{minipage}
    \begin{minipage}{0.34\linewidth}
    \caption{\small Depiction of the singular values $\mSigma_{Z}$ of the representation $\mZ$ learned by the BarlowTwins loss with $\mZ_{\rm a},\mZ_{\rm b}$ formed from \cref{eq:pair_form} given $\mG$ with $\rank(\mG)\in\{32,128\}$ ({\bf rows, green dotted lines}) and with $N$=$512,K$=$128$ during training ({\bf from blue to red}, number in top-right corner) with various initialization of $\mZ$ ({\bf columns}). We observe that the rank of the learned representation matches exactly the one of $\mG$  validating the result from \cref{thm:BT} regardless of the chosen hyper-parameters, making BarlowTwins's performances more sensitive to the design of $\mG$ akin to SimCLR and opposite to VICReg whose representation preserves full-rank (recall \cref{fig:VICReg_landscape2}) when tuning the invariance parameter.}
    \label{fig:BT}
    \end{minipage}
\end{figure}

The above statement also brings yet another flavor of SSL methods. In fact, where VICReg allows to control the rank of $\mZ$ to be in-between $K$ and $\rank(\mG)$ through the loss hyper-parameters, BarlowTwins and SimCLR enforce the rank of $\mZ$ to be exactly the rank of $\mG$. We depict in \cref{fig:BT} the evolution of $\rank(\mZ)$ depending on the rank of the initialized representation $\rank(\mZ_{\rm init})$ using a gradient descent optimizer. We see that regardless of the initial rank of $\mZ_{\rm init}$, training the representation to minimize the BarlowTwins loss makes the representation dimension collapse. Lastly, although not further studied here, we should point out to the reader that regularized forms of KCCA can be shown to include kernel ridge regression and regularized kernel Fisher LDA as special cases \citep{kuss2003geometry}, further tying the special cases for which different SSL methods would fall back to the same model.

In the following section we will demonstrate how BarlowTwins in the linear regime exactly recovers Canonical Correlation Analysis.


\subsection{With a Linear Network BarlowTwins Recovers Canonical Correlation Analysis and Linear Discriminant Analysis}
\label{sec:barlowtwins_linear}

The goal of this section is to further demonstrate the benefits of connecting SSL methods to spectral methods by exploiting the known techniques of the latter to help answer questions on the former.

As was done VICReg (we use linear settings of \cref{sec:vicreg_le}), we now obtain the optimal weights for BarlowTwins in the linear regime. We can even provide additional insights in this case since BarlowTwins is often seen as a key method that allows the use of different parameters/architectures to process $\mX_{\rm a}$ and $\mX_{\rm b}$. We now show under what conditions on $\mG$ sharing parameters is sufficient by first demonstrating how BarlowTwins recovers exactly CCA, and even LDA for supervised $\mG$. To streamline notations, we assume that our data is already centered, and thus define the covariance and cross-covariance matrices as $\mC_{\rm aa}=\mX_{\rm a}^T\mX_{\rm b},\mC_{lr}=\mX^T_{\rm a}\mX_{\rm b}$ and so on.

\begin{theorem}
\label{thm:BT_cca}
In the linear regime BarlowTwins recovers CCA with optimal weights  given by
\begin{align*}
    \mW^*_{\rm a}=\text{ top-$K$ eigenvectors of }\;\;\mC_{\rm aa}^{-1}\mC_{\rm ab}\mC_{\rm bb}^{-1}\mC_{\rm ba} \text{ and }\;\mW^*_{\rm b}=\mC_{\rm bb}^{-1}\mC_{\rm ba}\mW^*_{\rm a},
\end{align*}
and (i) ---with a symmetric $\mG$ and same dimensional $\mX_{a},\mX_{b}$, weight-sharing naturally occurs--- as the optimal weights are
$\mW^*_{\rm a}=\mW^*_{\rm b}=\text{ top-$K$ eigenvectors of }\;\;\mC_{\rm bb}^{-1}\mC_{\rm ba}$
and (ii) if $\mG$ is supervised and $K=C$ then BarlowTwins recovers LDA and thus VICReg (recall \cref{thm:vicreg_linear}). (Proof in \cref{proof:BT_cca}.)
\end{theorem}

\begin{lstlisting}[language=Python,escapechar=\%]
Caa = Xa.t() @ Xa + torch.eye(Da) * gamma # %\lsComment{\scriptsize$\mX_a \in \mathbb{R}^{N,D_a}$}% regularized with gamma %\lsComment{$(\gamma)$}%
Cbb = Xb.t() @ Xb + torch.eye(Db) * gamma # %\lsComment{\scriptsize$\mX_b \in \mathbb{R}^{N,D_b}$}% regularized with gamma %\lsComment{$(\gamma)$}%
RH = torch.block_diag(Caa, Cbb) # preparing the generalized eigenvalue prob.
LH = torch.block_diag(Xa.t() @ Xb, Xb.t() @ Xa).roll(Da, 1)
_, eigvecs = torch.lobpcg(LH, k=K, B=RH, niter=-1) # fast iterative solver
Wa, Wb = eigvecs.split([Da, Db])
Za = Xa @ Wa # %\lsComment{\scriptsize$\mZ_{\rm a} \in \mathbb{R}^{N,K}$}%
Zb = Xb @ Wb # %\lsComment{\scriptsize$\mZ_{\rm b} \in \mathbb{R}^{N,K}$}%
\end{lstlisting}

The above result opens new venues to extend current SSL methods (BarlowTwins in this case). For example, penalized matrix decomposition (PMD) from \citet{witten2009penalized} formulates a novel sparse formulation of CCA. In our context, this could lead to a new variation of BarlowTwins, in both the linear and nonlinear regimes. 
With the above results, we now connected most SSL methods to spectral methods, and found key properties that their representations/parameters inherit.

We now move away from connecting SSL methods to spectral methods, and finding the properties that their representations/parameters would inherit, to instead exploit those results and understand how well those learned representations can be to solve downstream tasks.

\section{Optimality of Self-Supervised Methods to Solve Downstream Tasks}
\label{sec:optimal}

The goal of this section is to answer the following question: {\em given a task ---encoded as a target matrix--- $\mY\in \mathbb{R}^{N \times C}$, what are the sufficient statistics of $\mY$ that a representation $\mZ\in \mathbb{R}^{N \times K}$ must preserve to ensure that $\min_{\mW,\vb}\| \mY-\mZ\mW-\vb\|_{F}^2=0$}. The first \cref{sec:MSE} will derive the optimal linear parameters (possibly non-unique) that minimize that loss, and will highlight the spectral properties of $\mY$ that must be consistent in $\mZ$. From that, \cref{sec:conditions} will be able to provide necessary and sufficient conditions for $\mZ$ to be optimal ---in term of its left-singular vectors--- and finally, \cref{sec:same_coin} extends the recent studies of \citet{bao2021sharp,haochen2021provable,haochen2022beyond} to multiple SSL methods by demonstrating how and when SSL representations are optimal for downstream tasks. 

\subsection{Characterizing a Representation Usefulness by its Ability to Linearly Solve a Task}
\label{sec:MSE}

The goal of this section is to characterize the (possibly non-unique) parameter $\mW$ of the linear transformation that given any target matrix $\mY$ and representation $\mZ$ minimize the Mean-Squared Error (MSE). Results in this section are standard in the linear algebra literature e.g. see \citet{golub1971singular} but we include it for completeness. We should also highlight that although for classification the cross-entropy loss is more common, it has recently been showed that $\ell_2$ could perform as well \citep{https://doi.org/10.48550/arxiv.2006.07322} with the correct parameter tuning, even on large models e.g. on the Imagenet dataset with modern deep learning architectures. Hence, we will leverage that loss function throughout this section.

Hence we consider that we are given representations $\vz\triangleq f(\vx)$ from an input sample $\vx$. Hence, we will denote the dataset representation as $\mZ=[f(\vx_1),\dots,f(\vx_N)]^T\in\mathbb{R}^{N \times K}$ given a dataset $\sX\triangleq \{\vx_1,\dots,\vx_N\},N\in \mathbb{N}^*$. We are also given a task ---encoded as a target matrix--- $\mY\triangleq [\vy_1,\dots,\vy_N]^T \in \mathbb{R}^{N \times C}$ where each $\vy_n$ is associated to each input $\vx_n$. 

Without loss of generality and to lighten our derivations, we will omit the bias vector $\vb$, it can be learned as part of $\mW$ by adding $1$ to the features of $\vz_n$. Our loss function thus takes the following form $\mathcal{L}(\mW) = \frac{1}{2}\|\mY -\mZ \mW \|_{F}^2$. To minimize this loss function, we will first find what matrices $\mW$ make the gradient of $\mathcal{L}$ with respect to $\mW$ vanish. Hence, we need to find $\mW \in \mathbb{R}^{K \times C}$ such that
\begin{align*}
    \nabla_{\mW} \mathcal{L}=&\mathbf{0} \iff -\mZ^T(\mY-\mZ \mW)=\mathbf{0}\iff  \mZ^T\mY=\mZ^T\mZ\mW.
\end{align*}
If we assumed $\mZ$ to be full rank, we would directly recover the usual least square solution $\mW^*=(\mZ^T\mZ)^{-1}\mZ^T\mY$. However, we would like to (i) avoid any assumption on the spectrum of $\mZ$, and (ii) avoid the use of any standard regularizer such as Tikhonov that is commonly used to recover a unique solution to an ill-posed optimization problem. In fact, our goal is to find the (possibly non-unique) family of parameters $\mW^*$ that fulfill $\nabla_{\mW} \mathcal{L}=\mathbf{0}$ regardless of the properties of $\mZ$ and $\mY$.
To that end, we first start by using the SVD of $\mZ=U_{z}\mSigma_z V_z^{\top}$ ---which always exists--- to reformulate the above equality (see derivations in \cref{proof:optimal_matrix,proof:optimal_value}) into
\begin{align}
    \mZ^T\mY=\mZ^T\mZ\mW \iff &\mW \in \left\{ (\mV_z\mSigma^{-1}_z\mU_{z}^T+\overline{\mV}_z\mM)\mY:\mM \in \mathbb{R}^{K \times C}\right\},\label{eq:solution}
\end{align}
with $\overline{\mV}_z$ the $K \times (K-\rank(\mZ))$ matrix that horizontally stacks the right singular vectors of $\mZ$ which have their corresponding singular value $0$, with the special case that $\mM=\mathbf{0} \iff \rank(\mZ)=K$. We also slightly abuse notations and define $\mSigma^{-1}_z$ to be the $K \times N$ matrix which is zero for all off-diagonal elements, and with 
$$(\mSigma^{-1}_z)_{i,j}\triangleq \begin{cases}
\frac{1}{(\sigma_z)_{i}} \iff i=j \wedge (\sigma_z)_{i} > 0\\
0 \text{ otherwise}
\end{cases},
$$
hence $\mSigma^{-1}_z\mSigma_z$ is a $K \times K$ matrix which is identity iff $\mX$ is full rank, and is otherwise filled with $K-\rank(\mX)$ zeros and $\rank(\mX)$ ones in the diagonal. On the other hand, $\mSigma_z\mSigma^{-1}_z$ is a diagonal $N \times N$ matrix with $N-\rank(\mX)$ zeros and $\rank(\mX)$ ones in the diagonal. Note that in the special case where $\mX$ is full-rank, $\overline{\mV}$ is null (we slightly abuse notations) and  $\mV_z\mSigma^{-1}_z\mU_{z}^T=(\mZ^T\mZ)^{-1}\mZ^T$ recovering the standard least-square solution.

\subsection{Necessary and Sufficient Conditions for Optimality of a Representation}
\label{sec:conditions}

Given a target matrix $\mY$ and its SVD $\mU_y\mSigma_y\mV^T_y$, we obtain the following formal statement that demonstrates how the left-singular vectors of $\mY$ must related to the left-singular vectors of $\mZ$ to allow for $\min_{\mW}\mathcal{L}(\mW)$ to be $0$. This statement plays a crucial role in our study as it answers the question {\em what property $\mZ$ must fulfill ---regardless on how it was produced--- to guarantee that $0$ training error is achievable on the considered task?}

\begin{theorem}[Necessary and sufficient condition]
\label{thm:optimal}
Given a task $\mY\in\mathbb{R}^{N\times C}$ and a representation $\mZ\in\mathbb{R}^{N\times K}$ ---with left-singular vectors associated to nonzero singular values denoted as $\widehat{\mU}_z$--- the minimum linear loss is given by 
\begin{align}
    \min_{\mW\in\mathbb{R}^{K\times C}}\mathcal{L}(\mW)=\frac{1}{2}\|\mY \|_{F}^2-\frac{1}{2}\|\widehat{\mU}_{z}^T\mU_{y}\mSigma_y \|_{F}^2,\label{eq:minimal}
\end{align}
which is $0$ iff the columns of $\widehat{\mU}_{z}$ spans the columns of $\widehat{\mU}_y$.
(Proof in \cref{proof:loss_value}.)
\end{theorem}

\begin{figure}[t!]
    \centering
    \begin{minipage}{0.7\linewidth}
    \includegraphics[width=\linewidth]{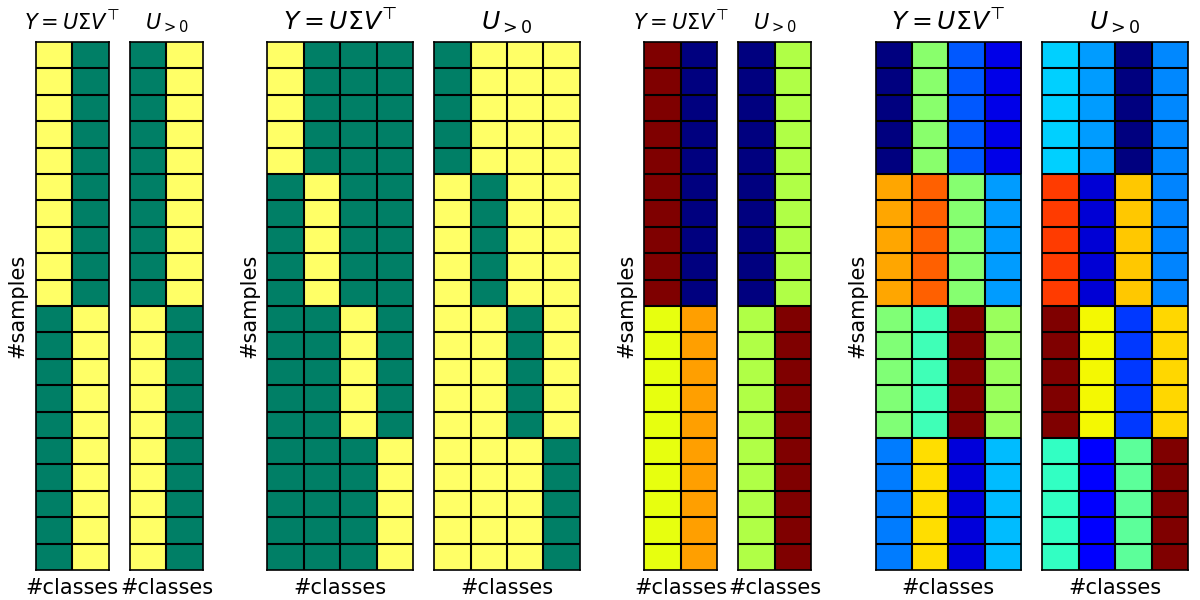}
    \end{minipage}
    \begin{minipage}{0.29\linewidth}
    \caption{\small Depiction of typical target matrices $\mY$  and their corresponding left singular vectors. Note that in this case all the nonzero singular values are identical, but with class-imbalance the singular values would be proportional to the class proportions. Hence it is clear that approximating the left singular vectors of $\mY$ via SSL training (recall \cref{thm:optimal}) the learned representations will exhibit per-class clustering ---as long as the left-singular vectors of $\mG$ correctly encode the task.}
    \label{fig:target_matrix}
    \end{minipage}
\end{figure}

The proof consists in using the solution $\mW$ from the set in \cref{eq:solution}, and after a few algebraic manipulations, \cref{thm:optimal} result is obtained. From that result alone we already obtain an interesting requirement on $\mZ$, namely that its top-$K$ left singular vectors must be the same ---up to a rotation--- to the top-$K$ left singular of $\mY$. \cref{thm:optimal} also provides us with a direct necessary condition but not sufficient condition for optimality.

Understanding the inter-play between $\rank(\mZ)$, $\mU_z$ and $\mU_y$ will play a crucial role in the next section where we propose to study self-supervised learning criterion, and their ability to produce optimal representations.

\subsection{Contrastive and Non-Contrastive Learning can all be Optimal}
\label{sec:same_coin}


We now demonstrate in this section that any representation $\mZ$ learned by any of the SSL method (VICReg, BarlowTwins, SimCLR) can be optimal for a downstream task, as long as the data geometry encoded in $\mG$ follows the left-singular vectors of $\mY$, the target matrix which embodies the considered downstream task.


\begin{theorem}
\label{thm:optimal}
Given a dataset $\mX$ and relation matrix $\mG$, minimizing the VICReg -or- SimCLR -or- BarlowTwins loss produces a representation that is optimal for a task $\mY=\mU_y\mSigma_y\mV_y^T$ iff the columns of $\widehat{\mU}_y$ are in the span of $\widehat{\mU}_{g}$ as in
\begin{align*}
\min_{\mW \in \mathbb{R}^{K \times C}} \|\mY -\mZ^*\mW \|_F^2 = 0 \iff \widehat{\mU}_y \in \spann (\widehat{\mU}_{g}),
\end{align*}
with $\mZ^*$ the embeddings of the VICReg -or- SimCLR -or- BarlowTwins model after convergence. (Proof in \cref{proof:optimal_vicreg}.)
\end{theorem}

Although not explicitly stated in \cref{thm:optimal} the same applies e.g. to NNCLR and MeanShift as they employ the same SimCLR loss, only the design of $\mG$ is altered. The above is crucial in helping and guiding the design of SSL methods and theoretically confirm the empirical findings from \citet{geirhos2020surprising} that observed in different scenarios that SSL and supervised models nearly fall back to the same thing.

\subsection{Non-Contrastive Methods Should be Preferred: Best and Worst Downstream Task Error Bounds}
\label{sec:not_equal}

We demonstrated in \cref{sec:same_coin} that all SSL methods can be optimal to solve a task at ahdn as long as the spectral properties of $\mG$ and $\mY$ are aligned. However, this is rarely the case in practical scenarios, and it thus becomes crucial to understand the behavior of the learned representation $\mZ$ on downstream task and if it varies with different SSL methods. First, we propose the following bound which represent the best and worst case downstream performances as a function of the rank of $\mG$ which is mostly a result of applying the Eckart-Young-Mirsky theorem. That is, we look at all the possible similarity matrices $\mG$ of rank $R$ and see, given a task $\mY$ what is the best achievable performance if $\mG$ correctly encoders the data geometry, and what is the worse possible performance if $\mG$ is ``orthogonal'' to the correct data geometry. For clarity and without loss of generality we assume here that $\rank(\mY)\leq K$ as otherwise no method would produce an optimal representation in general and $K <N$ as otherwise we are in the kernel regime.

\begin{theorem}
\label{cor:rank}
Given fixed inputs $\mX$ the lower and upper-bound over all possible matrices $\mG$ of rank $R$ of the downstream task training performances (with fixed $\mY$) are given by
\begin{align*}
\sum_{i=R}^{K}(\mSigma_y^2)_{i,i}\leq\left(\min_{\mW\in\mathbb{R}^{K\times C}}\mathcal{L}(\mW,\mZ^*_{\rm SimCLR}(\mG))-\min_{\mW\in\mathbb{R}^{K\times C}}\mathcal{L}(\mW,\mZ^*_{\rm VICReg}(\mG)) \right)\leq\|\mY\|_F^2,
\end{align*}
and are tight. Hence one should prefer VICReg, then BarlowTwins and finally SimCLR to maximize the downstream task performances.
\end{theorem}

The above result is a direct consequence of SimCLR forcing the representation to have the same rank as $\mG$ while VICReg always enforce a full-rank representation. And although this difference becomes irrelevant with correct $\mG$ (recall \cref{sec:same_coin}) it becomes an important distinctive attribute between SSL methods when $\mG$ is not optimal, which concerns most practical scenarios.
\section{Conclusions}

We provided in this study a unifying analysis of the major self-supervised learning methods covering VICReg (\cref{sec:vicreg}), SimCLR (\cref{sec:simclr}) and BarlowTwins (\cref{sec:BT}). In doing so, we were able to not only tie each of those methods and their variants to common spectral embedding techniques, but we were also able to find the commonalities between all those methods. Among the many insights that we obtained, the most crucial one is that whenever the similarity matrix $\mG$ is correctly defined with respect to a downstream task, any of those methods will produce an ideal representation that will perfectly solve the task at hand. In short, there is no benefit of one method versus any other. In the more realistic regime where $\mG$ might be misaligned with the downstream task, VICReg with lower invariance regularization hyper-parameter should be preferred. In that regime, the representation will include all the information from $\mG$ while preserving full-rank and thus allowing for the representation to be usable for other downstream task that are not encoded within $\mG$. This is in contrast with BarlowTwins and SimCLR that collapse the representation to embed all information about $\mG$ and nothing else.

At a more general level, we were able to parallel the contrastive versus non-contrastive dichotomy in SSL to the global versus local methods in spectral methods respectively. This led to further highlights into the strengths and weakness of each. For example, global approaches (contrastive SSL) tend to give a more faithful representation of the data’s global structure as the embedding aims to be metric-preserving. On the other hand, the local approaches (non-contrastive) provide useful embedding on a broader range of manifolds, whose local geometry is close to Euclidean, but whose global geometry may not be \citep{silva2002global}.

Beyond those results, we hope that the ties provided in this paper will stem a plurality of future work. One example would be to leverage the connection between SSL methods and spectral embedding methods to port existing results and techniques from one field to the other. In fact, such spectral methods have fallen short when dealing with high-dimensional datasets such as Imagenet. However, SSL methods have risen to become state-of-the-art on those datasets. The only difference between those lies in how $\mG$ is constructed.

\section*{Acknowledgements}

We thank Prof.~Pascal Vincent and Prof.~Surya Ganguli for providing key discussions, insights and related work references that played a key role in making this study as complete and self-contained as possible.

\bibliography{iclr2022_conference}
\bibliographystyle{plainnat}


\appendix

\newpage

\begin{center}
    \Huge
    Supplementary Materials
\end{center}

The supplementary materials is providing the proofs of the main's paper formal results. We also provide as much background results and references as possible throughout to ensure that all the derivations are self-contained.
Some of the below derivation do not belong to formal statements but are included to help the curious readers get additional insights into current SSL methods.

\section{Formal Statements Proofs}
\label{proof:section}

\subsection{VICReg Variance+Covariance Versus Representation's Singular Values}
\label{proof:var_cov}

This simple derivation demonstrates how minimizing the VCReg can be done through an upper bound by constraining all the singular-values to be close to $1$ although the general criterion only enforces for the variance term to be ---at least--- 1 through the following derivations
\begin{align*}
    \min_{\mZ}\min_{\vu \in [1,\infty)^{K}}\|\mZ^T\mZ-\diag(\vu) \|_F^2&=\min_{\mZ}\min_{\vu \in [1,\infty)^{K}}\| \mV_{\mZ}\mSigma^2_{\mZ}\mV_{\mZ}^T-\diag(\vu)\|_F^2\\
    &=\min_{\mZ}\min_{\vu \in [1,\infty)^{K}}\|\mSigma^2_{\mZ}-\diag(\vu)\|_F^2\\
    &=\min_{\mZ}\min_{\vu \in [1,\infty)^{K}}\|\vsigma^2_{\mZ}-\vu\|_2^2\\
    &\leq\min_{\mZ}\|\vsigma^2_{\mZ}-\mathbf{1}\|_2^2
\end{align*}
where we denoted by $\vsigma_{\mZ}$ the diagonal part of the diagonal $\mSigma_{\mZ}$ matrix.

\subsection{Non-Unique Solution to Least-Square}
\label{proof:optimal_matrix}
The below derivation demonstrates that even in the representation $\mZ$ (or any input matrix) is not full-rank, the least-square type of solution $\mW$ to predict $\mY=\mZ\mW$ can be found, it is just not unique. In fact, it is possible to find an entire space of matrices that will minimize the loss i.e. respect the below equality that makes the loss have a gradient of $0$ and be at its minimum value, simply by moving within the kernel space of the representation (input) matrix as in
\begin{align*}
    \mZ^T\mY=\mZ^T\mZ\mW\iff &(U_{\mZ}\Sigma_{\mZ} V_{\mZ}^{\top})^T\mY=(U_{\mZ}\Sigma_{\mZ} V_{\mZ}^{\top})^\top(U_{\mZ}\Sigma_{\mZ} V_{\mZ}^{\top})\mW\\
    \iff &V_{\mZ}\Sigma_{\mZ}^TU_{\mZ}^T\mY=V_{\mZ}\Sigma_{\mZ}^T \Sigma_{\mZ} V_{\mZ}^{\top}\mW\\
    \iff &\mW^* \in \{ (V_{\mZ}\Sigma^{-1}_{\mZ}U_{\mZ}^T+\mM)\mY:\mM \in \ker(\mZ)\},
\end{align*}
where we recall that $\ker(\mZ)$ is the kernel space of$\mZ$, and where we slightly abused notations by employing $\Sigma^{-1}_{\mZ}$ to represent the inverse only for the non-zero element of the diagonal matrix $\Sigma_{\mZ}$ (there are as many zeros in the diagonal as the dimension of $\ker(\mZ)$).

\subsection[Proof]{Any Linear Weight from \cref{proof:optimal_matrix} Has Zero Least-Square Gradient}
\label{proof:optimal_value}

This section continues the previous derivations but now demonstrating that using this optimal value for $\mW$ denoted as $\mW^*$ does fullfil the equality i.e. we are at a global optimum for any matrix within the defined subspace as in
\begin{align*}
    V_{\mZ}\Sigma_{\mZ}^TU_{\mZ}^T\mY=&V_{\mZ}\Sigma_{\mZ}^T \Sigma_{\mZ} V_{\mZ}^{\top}\mW^*\\
    \implies  V_{\mZ}\Sigma_{\mZ}^TU_{\mZ}^T\mY=&V_{\mZ}\Sigma_{\mZ}^T \Sigma_{\mZ} V_{\mZ}^{\top}(V_{\mZ}\Sigma^{-1}_{\mZ}U_{\mZ}^T+\mM)\mY\\
    \implies  V_{\mZ}\Sigma_{\mZ}^TU_{\mZ}^T\mY=&V_{\mZ}\Sigma_{\mZ}^T \Sigma_{\mZ} \Sigma^{-1}_{\mZ}U_{\mZ}^T\mY\\
    \implies  V_{\mZ}\Sigma_{\mZ}^TU_{\mZ}^T\mY=&V_{\mZ}\diag(\sigma_{\mZ})^2 \Sigma^{-1}_{\mZ}U_{\mZ}^T\mY\\
    \implies  V_{\mZ}\Sigma_{\mZ}^TU_{\mZ}^T\mY=&V_{\mZ}\Sigma_{\mZ}^TU_{\mZ}^T\mY
\end{align*}
where the last equality follows since $\diag(\sigma_{\mZ})^2 \Sigma^{-1}_{\mZ}$ will either multiply $(\sigma_{\mZ})_i^2$ with $(\Sigma^{-1}_{\mZ})_{i,i}=\frac{1}{(\sigma_{\mZ})_i}$ if $(\sigma_{\mZ})_i$ is nonzero, and otherwise if $(\sigma_{\mZ})_i$ is $0$, then it will be the product between $0$ for $(\sigma_{\mZ})_i^2$ and $0$ for $(\Sigma^{-1}_{\mZ})_{i,i}$ giving back the original value for $(\sigma_{\mZ})_i$.

\subsection{Achievable Loss with Low-Rank Representation}
\label{proof:loss_value}
This section takes a last detour towards understanding the least-square loss with a low-rank input/representation matrix. In this case we derive a various set of quantities that quantify the minimum loss that is achieved by any of the optimal matrix $\mW^*$ found in \cref{proof:optimal_matrix} as follows again slightly abusing notations for $\Sigma^{-1}_{\mZ}$ to only invert the non-zero singular values of $\mZ$ as
\begin{align*}
    \frac{1}{2}\|\mY -\mZ \mW^* \|_{F}^2=&\frac{1}{2}\|\mY -\mZ (V_{\mZ}\Sigma^{-1}_{\mZ}\mU_{\mZ}^T+\tilde{V}_{\mZ}\mM)\mY \|_{F}^2\\
    =&\frac{1}{2}\|\mY -\mU_{\mZ}\Sigma_{\mZ} \mV_{\mZ}^{\top} \mV_{\mZ}\Sigma^{-1}_{\mZ}\mU_{\mZ}^T\mY \|_{F}^2\\
    =&\frac{1}{2}\|\mY -\mU_{\mZ}\Sigma_{\mZ}\Sigma^{-1}_{\mZ}\mU_{\mZ}^T\mY \|_{F}^2\\
    =&\frac{1}{2}\|(\mI -\mU_{\mZ}\Sigma_{\mZ}\Sigma^{-1}_{\mZ}\mU_{\mZ}^T)\mY \|_{F}^2\\
    =&\frac{1}{2}\|(\mU_{\mZ}\mU_{\mZ}^T -\mU_{\mZ}\Sigma_{\mZ}\Sigma^{-1}_{\mZ}\mU_{\mZ}^T)\mY \|_{F}^2\\
    =&\frac{1}{2}\|(\mI -\Sigma_{\mZ}\Sigma^{-1}_{\mZ})U_{\mZ}^T\mY \|_{F}^2\\
    =&\frac{1}{2}\|\mY \|_{F}^2-\frac{1}{2}\|\Sigma_{\mZ}\Sigma^{-1}_{\mZ}\mU_{\mZ}^T\mY \|_{F}^2\\
    =&\frac{1}{2}\|\mY \|_{F}^2-\frac{1}{2}\|\Sigma_{\mZ}\Sigma^{-1}_{\mZ}U_{\mZ}^T\mU_{y}\Sigma_y \|_{F}^2\\
    =&\frac{1}{2}\|\mY \|_{F}^2-\frac{1}{2}\sum_{i,j}1_{\{(\sigma_{\mZ})_i>0\}}\langle (\mU_{\mZ})_i,(\mU_y)_j\rangle^2(\sigma_y)^2_j,
\end{align*}
where it is clear that the minimum loss will in general not be $0$ unless $\Sigma_{\mZ}\Sigma^{-1}_{\mZ}=\mI$ which is not guaranteed (recall that we abuse notation for the inverse and that $\Sigma_{\mZ}\Sigma^{-1}_{\mZ}$ is only $1$ in its diagonal for the nonzero singular values of $\mZ$.

\subsection[Proof]{Equivalence Between VICReg Invariance Term and Trace with Graph Laplacian}
\label{proof:invariance_trace}

The goal of this section is to derive the first crucial result of our study that ties VICReg to spectral embedding methods by doing a first connection between VICReg invariance loss and the Dirichlet energy of a graph.
The equality follows from the Laplacian of the graph definition $\mL=\mD-\mG$ where $\mD$ is the degree matrix of the graph i.e. a diagonal matrix with entries corresponding to the sum of each row of $\mG$, and using the following algebraic manipulations which are common in the spectral graph analysis community, see e.g. \citep{von2007tutorial}
\begin{align*}
    \Tr\left(\mZ^T\mL\mZ\right)=&\sum_{d}(\mZ^T\mL\mZ)_{d,d}=\sum_{i,j}\mL_{i,j}\langle (\mZ)_{j,.}.(\mZ)_{i,.}\rangle\\
    =&\sum_{i}\|(\mZ)_{i,.} \|_2^2(\mD)_{i,i}-\sum_{i,j}(\mG)_{i,j}\langle (\mZ)_{j,.}.(\mZ)_{i,.}\rangle\\
    =&\frac{\sum_{i}\|(\mZ)_{i,.} \|_2^2(\mD)_{i,i}+\sum_{i}\|(\mZ)_{i,.} \|_2^2(\mD)_{i,i}}{2}-\sum_{i,j}(\mG)_{i,j}\langle (\mZ)_{j,.}.(\mZ)_{i,.}\rangle\\
    =&\frac{\sum_{i}\sum_{j}\|(\mZ)_{i,.} \|_2^2(\mG)_{i,j}+\sum_{i}\sum_{j}\|(\mZ)_{i,.} \|_2^2(\mG)_{i,j}}{2}-\sum_{i,j}(\mG)_{i,j}\langle (\mZ)_{j,.},(\mZ)_{i,.}\rangle\\
    =&\frac{1}{2}\sum_{i}\sum_{j}(\mG)_{i,j}\|(\mZ)_{i,.}-(\mZ)_{j,.}\|_2^2,
\end{align*}
which is a famous derivation in graph signal processing relating pairwise distances to Dirichlet energy of the underlying graph with Laplacian $\mL$.

\subsection{Non-Uniqueness of a Representation to a Given VICReg Loss Value}
\label{proof:not_unique}

In this section we provide a simple argument to demonstrate that the VICReg representation that obtains a loss value of $c$ is not unique, regardless of the (achievable) value of $c$. To see that, one can for example add a constant vector to each row of $\mZ$ and see that \cref{eq:VICReg} is left unchanged. In fact the computation of the covariance matrix is invariant to constant column shift of $\mZ$, and the invariance term will automatically cancel those added vectors when comparing pairs of rows. 

\subsection[Proof]{Optimal Representation and Loss for VICReg (\cref{thm:VICReg_optimal})}
\label{proof:VICReg_optimal}



The goal of this section is to obtain the closed-form optimal representation of VICReg using the least-square variance loss, instead of the hinge-loss, and to find the minimum loss associated to that (non-unique) optimum. Using the Trace term derivations given in \cref{proof:invariance_trace} we obtain
\begin{align*}
    \mathcal{L}_{\rm VIC}=\alpha\| \Cov(\mZ^*_{\alpha,\gamma})-\mI\|_F ^2+2\frac{\gamma}{N}\Tr\left((\mP'_{\alpha,\gamma}(\mLambda'_{\alpha,\gamma} N)^{1/2})^T\mL\mP'_{\alpha,\gamma}(\mLambda'_{\alpha,\gamma} N)^{1/2}\right),
\end{align*}
where we use $\vlambda_{\alpha,\beta}',\mP'_{\alpha,\gamma}$ and $\mLambda'_{\alpha,\gamma}$ to denote the first $K$ indices/columns of \cref{eq:VICReg_optimal}. Simplifying the trace term leads to
\begin{align*}
    \frac{\gamma}{N}\Tr\left((\mP'_{\alpha,\gamma}(\mLambda'_{\alpha,\gamma} N)^{1/2})^T\mL\mP'_{\alpha,\gamma}(\mLambda'_{\alpha,\gamma} N)^{1/2}\right)&=
    \gamma\Tr\left(\mLambda'_{\alpha,\gamma} {\mP'}_{\alpha,\gamma}^T\mL\mP'_{\alpha,\gamma}\right)\\
    &=\alpha\Tr\left(\mLambda'_{\alpha,\gamma} {\mP'}_{\alpha,\gamma}^T\mM\mP'_{\alpha,\gamma}\right)\\
    &\hspace{2.5cm}-\alpha\Tr\left(\mLambda'_{\alpha,\gamma} {\mP'}_{\alpha,\gamma}^T(\mM-\frac{\gamma}{\alpha}\mL)\mP'_{\alpha,\gamma}\right)\\
    &=\alpha\Tr\left(\mLambda'_{\alpha,\gamma} {\mP'}_{\alpha,\gamma}^T(\mI-\frac{1}{N}\mathbf{1}\mathbf{1}^T)\mP'_{\alpha,\gamma}\right)
    -\alpha\|\vlambda'_{\alpha,\gamma}\|_2^2,
\end{align*}
plugging this value into the loss we obtain
\begin{align*}
    \mathcal{L}_{\rm VIC}=&\alpha\left(\| \Cov(\mZ^*_{\alpha,\gamma})-\mI\|_F ^2+2\Tr\left(\mLambda'_{\alpha,\gamma} {\mP'}_{\alpha,\gamma}^T(\mI-\frac{1}{N}\mathbf{1}\mathbf{1}^T)\mP'_{\alpha,\gamma}\right)
    -2\|\vlambda'_{\alpha,\gamma}\|_2^2\right)\\
    =&\alpha\left(\| \Cov(\mZ^*_{\alpha,\gamma})\|_F^2+K-2\Tr(\Cov(\mZ^*_{\alpha,\gamma})) +2\Tr\left(\mLambda'_{\alpha,\gamma} {\mP'}_{\alpha,\gamma}^T(\mI-\frac{1}{N}\mathbf{1}\mathbf{1}^T)\mP'_{\alpha,\gamma}\right)
    -2\|\vlambda'_{\alpha,\gamma}\|_2^2\right)\\
    =&\alpha\left(\| \Cov(\mZ^*_{\alpha,\gamma})\|_F^2+K
    -2\|\vlambda'_{\alpha,\gamma}\|_2^2\right)\\
    =&\alpha\left(\left\| \frac{1}{N}(\mZ^*_{\alpha,\gamma})^T\mM\mZ^*_{\alpha,\gamma}\right\|_F^2+K
    -2\|\vlambda'_{\alpha,\gamma}\|_2^2\right),
\end{align*}
now one should recall that $\mZ^*_{\alpha,\gamma}$ contains the eigenvectors of $\mM-\frac{\gamma}{\alpha}\mL$ and that each of those eigenvector that has nonzero singular value has $0$ mean since 
\begin{align*}
    (\mM-\frac{\gamma}{\alpha}\mL)\vv&=\lambda \vv \\
    \iff \mM (\mI-\frac{\gamma}{\alpha}\mL)\mM\vv&=\lambda \vv&&\text{(Laplacian rows/cols sum to $0$)}\\
    \implies \mathbf{1}^T\mM (\mI-\frac{\gamma}{\alpha}\mL)\mM\vv&=\lambda \mathbf{1}^T\vv\\
    \implies 0&=\lambda \mathbf{1}^T\vv \implies \mathbf{1}^T\vv=0&&\text{(for any eigenvector $\vv$ with $\lambda>0$)}
\end{align*}
hence we obtain the following simplifications
\begin{align*}
    \mathcal{L}_{\rm VIC}=&\alpha\left(\left\| \frac{1}{N}(\mZ^*_{\alpha,\gamma})^T\mM\mZ^*_{\alpha,\gamma}\right\|_F^2+K
    -2\|\vlambda'_{\alpha,\gamma}\|_2^2\right)\\
    =&\alpha\left(\left\| \frac{1}{N}(\mZ^*_{\alpha,\gamma})^T\mZ^*_{\alpha,\gamma}\right\|_F^2+K
    -2\|\vlambda'_{\alpha,\gamma}\|_2^2\right)\\
    =&\alpha\left(\left\| \vlambda'_{\alpha,\gamma}\right\|_2^2+K
    -2\|\vlambda'_{\alpha,\gamma}\|_2^2\right)\\
    =&\alpha (K-\|\vlambda'_{\alpha,\gamma}\|_2^2)
\end{align*}
which conludes the proof.

\subsection[Proof]{Proof of VICReg Recovering Laplacian Eigenmaps (\cref{thm:LE})}
\label{proof:vicreg_le}

This section takes on proving the first key result of our study that thoroughly tie VICReg to the known local spectral embedding method Laplacian Eigenmap. The only difference would be that in our case the graph $\mG$ is given from an SSL viewpoint and not constructed from a k-NN graph and geodesic distance estimate as in LE. We already know from \cref{proof:invariance_trace} that the invariance term of VICReg corresponds to the Trace term (times $2$) that LE tries to minimize. The only thing that we have to show to have the equivalent between the two is that the constraint $\mZ^T\mD\mZ=\mI$ is equivalent to the one that imposes an exact minimization of the variance and covariance terms. To see that, first notice (or recall from \cref{proof:VICReg_optimal}) that the variance+covariance term can be expressed as $
    \| \frac{1}{N}\mZ^T\mM\mZ-\mI\|_F^2$. 
Now, recall that the minimizer of LE consists in taking the $[2:K+1]$ eigenvectors of the Laplacian matrix $\mI-\mD^{-1}\mG$ which is equivalent (up to a rescaling of the eigenvalues) to taking the eigenvectors of $\mD-\mG$ as long as $\mD$ is isotropic. We also already saw that those eigenvectors either have zero mean and nonzero eigenvalue, or have arbitrary means and zero eigenvalue. As LE only considers the eigenvectors with nonzero eigenvalues, it is direct to see that those eigenvectors are centered. This translates to $
    \| \frac{1}{N}\mZ^T\mM\mZ-\mI\|_F^2=\| \frac{1}{N}\mZ^T\mZ-\mI\|_F^2$.
Hence, up to a rescaling, as long as $D$ is isotropic ($D=c\mI$) which is always the case in SSL, enforcing $\mathcal{L}_{\rm var}=0,\mathcal{\rm cov}=0$ is equivalent to $\mZ^T\mZ=\frac{N}{c}\mI$.

\subsection[Proof]{Proof of Linear VICReg Optimal Parameters (\cref{eq:Wstar_vicreg})}
\label{proof:linear_vicreg}

We are again using the quadratic variance term in-place of the hinge-term at $1$. In that setting, the linear VICReg loss falls back to
\begin{align*}
    \mathcal{L}_{\rm vic} = \alpha \left\| \frac{1}{N}\mW^T\mX^T \mH \mX \mW - \mI\right\|_F^2 +   \frac{2\gamma}{N}\Tr(\mW^T\mX^T\mL\mX\mW),
\end{align*}
which can be written in term of traces only to simplify differentiation as
\begin{align*}
    \mathcal{L}_{\rm vic} =& \frac{\alpha}{N^2}\Tr( \mW^T\mX^T \mH \mX \mW\mW^T\mX^T \mH \mX \mW)\\
    &-\frac{2\alpha}{N}\Tr(\mW^T\mX^T \mH \mX \mW) + \frac{2\gamma}{N} \Tr(\mW^T\mX^T\mL\mX\mW)+{\rm cst}\\
    =&\frac{\alpha}{N^2}\Tr( \mW^T\mX^T \mH \mX \mW\mW^T\mX^T \mH \mX \mW)+\frac{2}{N}\Tr(\mW^T\mX^T (\gamma \mL-\alpha\mH) \mX \mW)+{\rm cst}
\end{align*}
which we can now differentiate with respect to the $\mW$ parameter to obtain
\begin{align*}
    \nabla_{\mW}\mathcal{L}_{\rm vic} = \frac{4\alpha}{N^2}\mX^T \mH \mX \mW\mW^T\mX^T \mH \mX \mW+ \frac{4}{N}\mX^T (\gamma \mL-\alpha\mH) \mX \mW,
\end{align*}
which we need to set to $0$ and thus can be simplified to (since $\alpha > 0$)
\begin{align*}
    \nabla_{\mW}\mathcal{L}_{\rm vic} = \frac{1}{N}\mX^T \mH \mX \mW\mW^T\mX^T \mH \mX \mW+ \mX^T \left(\frac{\gamma}{\alpha} \mL-\mH\right) \mX \mW.
\end{align*}
First, let's consider that $\mX^T \mH \mX$ is invertible i.e. the data lies on a $D$-dimensional affine space. If not, the original data $\mX$ can simply be projected onto its subspace prior applying VICReg. In that setting and if $K=D$, we directly obtain
\begin{align*}
    \nabla_{\mW}\mathcal{L}_{\rm vic} =& \mathbf{0}\\
    \iff \frac{1}{N}\mX^T \mH \mX \mW\mW^T\mX^T \mH \mX \mW=& \mX^T \left(\mH-\frac{\gamma}{\alpha} \mL\right) \mX \mW\\
    \iff \mW\mW^T\Cov(\mX) \mW =& \frac{1}{N}\Cov(\mX)^{-1}\mX^T \left(\mH-\frac{\gamma}{\alpha} \mL\right) \mX \mW,
\end{align*}
denoting for clarity $\mA\triangleq \Cov(\mX)$ and $\mB\triangleq \frac{1}{N}\mX^T \left(\mH-\frac{\gamma}{\alpha} \mL\right) \mX=\Cov(\mX)-\frac{\gamma}{\alpha}\mX^T \mL\mX$ we have
\begin{align*}
    \mW\mW^T\mA \mW =& \mA^{-1}\mB \mW,
\end{align*}
which is solved for $\mW=\mA^{-1}\sqrt{\mB}$. Now, if $K<D$ then there are multiple local minimum, all with the same loss value, and those are obtained by extracting any $K$-out-of$D$ columns of the above solution. For \cref{eq:Wstar_vicreg} we considered the first $K$ columns arbitrarily.

\subsection[Proof]{Proof of Laplacian estimation with contrastive learning \cref{lemma:infoNCE}}
\label{proof:contrastive}

The first step of the proof consists in recovering the softmax with any metric $d$ that computes the distance (whatever distance desired) between pairs of inputs.

{\bf Graph Laplacian Estimation recovers the step 1 of contrastive methods using $\mathcal{R}_{\rm log}$.}~To prove \cref{eq:opti_W_simclr} (the case with $\mathcal{G}$ and not $\mathcal{G}_{\rm rsto}$ can be done similarly be removing the row-sum-to-one constraint), we need to solve the following optimization problem
\begin{align*}
    \argmin_{\mW \in \mathcal{G}_{\rm rsto}}\sum_{i\not =j}d(f_{\theta}(\vx_i),f_{\theta}(\vx_j))\mW_{i,j}+\tau\sum_{i\not =j}\mW_{i,j} (\log(\mW_{i,j})-1),
\end{align*}
which we solve by introducing the constraint in the optimization problem with the Lagrangian to
\begin{align*}
    \mathcal{L}=&\sum_{i\not =j}d(f_{\theta}(\vx_i),f_{\theta}(\vx_j))\mW_{i,j}+\tau\sum_{i\not =j}\mW_{i,j} (\log(\mW_{i,j})-1)+\sum_{i=1}^{N}\lambda_i(\sum_{j\not = i}W_{i,j}-1)+\sum_{i}\beta_i\mW_{i,i}
\end{align*}
\begin{align*}
    \implies \frac{\partial \mathcal{L}}{\partial \mW_{i,j}}=&\begin{cases}
    d(f_{\theta}(\vx_i),f_{\theta}(\vx_j))+\tau\log(\mW_{i,j})+\lambda_i \iff i\not = j\\
    \beta_i \iff i=j
    \end{cases}
    \\
    \implies \frac{\partial \mathcal{L}}{\partial \lambda_{i}}=&\sum_{j\not = i}\mW_{i,j}-1\\
    \implies \frac{\partial \mathcal{L}}{\partial \beta{i}}=&\mW_{i,i},
\end{align*}
we first solve for $\frac{\partial \mathcal{L}}{\partial \mW_{i,j}}=0$ for $i\not = j$ to obtain an expression of $\mW_{i,j}$ as a function of $\lambda_i$ as
\begin{align*}
    \frac{\partial \mathcal{L}}{\partial \mW_{i,j}}=0
    \iff  & d(f_{\theta}(\vx_i),f_{\theta}(\vx_j))+\tau\log(\mW_{i,j})+\lambda_i=0\\
    \iff & \tau\log(\mW_{i,j})=-d(f_{\theta}(\vx_i),f_{\theta}(\vx_j))-\lambda_i\\
    \iff & \mW_{i,j}=e^{\frac{-1}{\tau}(d(f_{\theta}(\vx_i),f_{\theta}(\vx_j))+\lambda_i)}\\
    \iff & \mW_{i,j}=e^{\frac{-1}{\tau}d(f_{\theta}(\vx_i),f_{\theta}(\vx_j))}e^{\frac{-1}{\tau}\lambda_i},
\end{align*}
and now using $\frac{\partial \mathcal{L}}{\partial \lambda_{i}}=0$ we will be able to solve for $\lambda_i$ as follows
\begin{align*}
    \frac{\partial \mathcal{L}}{\partial \lambda_{i}}=0 &\iff \sum_{j\not = i}\mW_{i,j}-1=0\\
    &\iff e^{\frac{-\lambda_i}{\tau}}\sum_{j\not = i} e^{\frac{-1}{\tau}d(f_{\theta}(\vx_i),f_{\theta}(\vx_j))}=1\\
    &\iff e^{\frac{-\lambda_i}{\tau}}=\frac{1}{\sum_{j\not = i} e^{\frac{-1}{\tau}d(f_{\theta}(\vx_i),f_{\theta}(\vx_j))}},
\end{align*}
which allows us to finally obtain an explicit solution for $\frac{\partial \mathcal{L}}{\partial \mW_{i,j}}=0$ that does not depend on $\lambda_i$ as follows
\begin{align*}
    \frac{\partial \mathcal{L}}{\partial \mW_{i,j}}=0&\iff \mW_{i,j}=e^{\frac{-1}{\tau}d(f_{\theta}(\vx_i),f_{\theta}(\vx_j))}e^{\frac{-\lambda_i}{\tau}}\\
    &\iff \mW_{i,j}=\frac{e^{\frac{-1}{\tau}d(f_{\theta}(\vx_i),f_{\theta}(\vx_j))}}{\sum_{j\not = i} e^{\frac{-1}{\tau}d(f_{\theta}(\vx_i),f_{\theta}(\vx_j))}}
\end{align*}
which holds for $i \not = j$. Now for the case $i=j$ simply using the constraint and solving for $\beta_i$ directly gives $\mW_{i,i}=0$. Notice that we did not enforce the $\mW_{i,j}=\mW_{j,i}$ constraint, however, the optimum found fulfills it and thus we are in a scenario akin to undirected graph Laplacian estimation.

{\bf Recovering SimCLR with Cosine Similarity.}~Now, since we are using the cosine distance, defined as $f(\vx,\vy)=1-\frac{\langle \vx,\vy\rangle}{\|\vx\| \|\vy \|}$, we can plug it in the above to finally obtain
\begin{align*}
    \mW_{i,j}=\frac{e^{\frac{-1}{\tau}d(f_{\theta}(\vx_i),f_{\theta}(\vx_j))}}{\sum_{j\not = i} e^{\frac{-1}{\tau}d(f_{\theta}(\vx_i),f_{\theta}(\vx_j))}}1_{\{i \not = j\}}=\frac{e^{\frac{1}{\tau}\frac{\langle f_{\theta}(\vx_i),f_{\theta}(\vx_j)\rangle}{\|f_{\theta}(\vx_i)\|_2 \|f_{\theta}(\vx_j) \|_2}}}{\sum_{j\not = i} e^{\frac{1}{\tau}\frac{\langle f_{\theta}(\vx_i),f_{\theta}(\vx_j)\rangle}{\|f_{\theta}(\vx_i)\|_2 \|f_{\theta}(\vx_j) \|_2}}}1_{\{i \not = j\}},
\end{align*}
which is exactly the features used by SimCLR or NNCLR. The last step is direct, take the graph of known positive pairs of nearest neighbor, apply the cross entropy between those and the above, and one obtains that \cref{eq:opti_W_simclr} recovers exactly the loss of those models.

{\bf Graph Laplacian Estimation recovers the step 1 of contrastive methods using $\mathcal{R}_{\rm F}$.}~~We will be using the $\mathcal{G}_{\rm rsto}$ space again as the case for $\mathcal{G}$ can be obtained easily by removing the Lagrangian constraint. we need to solve the following optimization problem
\begin{align*}
    \argmin_{\mW \in \mathcal{G}_{\rm rsto}}\sum_{i\not =j}d(f_{\theta}(\vx_i),f_{\theta}(\vx_j))\mW_{i,j}+\tau\sum_{i\not =j}\mW_{i,j} (\mW_{i,j}/2-1),
\end{align*}
which is augmented with the constraints to
\begin{align*}
    \mathcal{L}=\sum_{i\not =j}d(f_{\theta}(\vx_i),f_{\theta}(\vx_j))\mW_{i,j}+\tau\sum_{i\not =j}\mW_{i,j} (\mW_{i,j}/2-1)+\sum_{i=1}^{N}\lambda_i(\sum_{j\not = i}W_{i,j}-1)+\sum_{i}\beta_i\mW_{i,i}
\end{align*}
which we will differentiate with respect to $\mW,\lambda,\beta$ to obtain given $\mathcal{L}$ above
\begin{align*}
    \implies \frac{\partial \mathcal{L}}{\partial \mW_{i,j}}=&\begin{cases}
    d(f_{\theta}(\vx_i),f_{\theta}(\vx_j))+\tau\mW_{i,j}-\tau+\lambda_i \iff i\not = j\\
    \beta_i \iff i=j
    \end{cases}
    \\
    \implies \frac{\partial \mathcal{L}}{\partial \lambda_{i}}=&\sum_{j\not = i}\mW_{i,j}-1\\
    \implies \frac{\partial \mathcal{L}}{\partial \beta{i}}=&\mW_{i,i},
\end{align*}
setting $\frac{\partial \mathcal{L}}{\partial \mW_{i,j}}$ to $0$ we will obtain the following simplification isolating $\lambda_i$ for $i\not = j$
\begin{align*}
    \frac{\partial \mathcal{L}}{\partial \mW_{i,j}}=0\iff & d(f_{\theta}(\vx_i),f_{\theta}(\vx_j))+\tau\mW_{i,j}-\tau+\lambda_i =0
    \\
    \iff &\mW_{i,j}=1-\frac{1}{\tau}\Big(d(f_{\theta}(\vx_i),f_{\theta}(\vx_j))+\lambda_i\Big)
\end{align*}
and now using $\frac{\partial \mathcal{L}}{\partial \lambda_{i}}=0$ we will obtain
\begin{align*}
    \frac{\partial \mathcal{L}}{\partial \lambda_{i}}=0 \iff & \sum_{j\not = i}\mW_{i,j}-1 = 0\\ 
    \iff & \sum_{j\not = i}\bigg(1-\frac{1}{\tau}\Big(d(f_{\theta}(\vx_i),f_{\theta}(\vx_j))+\lambda_i\Big) \bigg)-1 = 0\\ 
    \iff & N-1+\frac{-\sum_{j\not = i}d(f_{\theta}(\vx_i),f_{\theta}(\vx_j))}{\tau}-\frac{(N-1)}{\tau}\lambda_i-1 = 0\\
    \iff & \lambda_i=\frac{-\sum_{j\not = i}d(f_{\theta}(\vx_i),f_{\theta}(\vx_j))}{(N-1)}+\tau(1-\frac{1}{N-1})
\end{align*}
and then plugging that into the original system of equation leads to
\begin{align*}
    \frac{\partial \mathcal{L}}{\partial \mW_{i,j}}=0\iff &\mW_{i,j}=1-\frac{1}{\tau}\Big(d(f_{\theta}(\vx_i),f_{\theta}(\vx_j))+\lambda_i\Big)\\
    \iff & \mW_{i,j}=1-\frac{1}{\tau}\Bigg(d(f_{\theta}(\vx_i),f_{\theta}(\vx_j))+\Big(\frac{-\sum_{j\not = i}d(f_{\theta}(\vx_i),f_{\theta}(\vx_j))}{(N-1)}+\tau\big(1-\frac{1}{N-1}\big)\Big)\Bigg)\\
    \iff & \mW_{i,j}=\frac{1}{N-1}-\frac{1}{\tau}\Big(d(f_{\theta}(\vx_i),f_{\theta}(\vx_j))-\frac{\sum_{j\not = i}d(f_{\theta}(\vx_i),f_{\theta}(\vx_j))}{(N-1)}\Big)
\end{align*}
for any $i\not = j$. Again solving for $\beta_i$ will lead directly to $\mW_{i,i}=0$. In matrix form, we thus obtain $\mW = \frac{1}{N-1}(\mathbf{1}\mathbf{1}^T-\mI)-\frac{1}{\tau}\left(\mD-\frac{1}{N-1}\mD\mathbf{1}\mathbf{1}^T\right)$ since the diagonal elements of $\mD$ are $0$.



\subsection[Proof]{Proof of SimCLR \cref{thm:simclr}}
\label{proof:simclr}

The goal of this section is to demonstrate that SimCLR, although optimizing the graph estimate $\widehat{\mG}$ to match $\mG$ forces the representation $\mZ$ to match $\mG$ through its outer-product. This result closely follows a multidimensional scaling type of reasoning which should not be a surprise being a global spectral embedding method.
Let's say that $\tau$ is for example $1$ and that all distances are between $0$ and $1$ e.g. using the cosine distance to streamline the derivations (if not, simply set $\tau$ to the maximum value present in the matrix $\mD$). We thus obtain
\begin{align*}
    \| \mG -\widehat{\mG}\|_F^2=\| (\mathbf{1}\mathbf{1}^T-\mI-\mG) -((\mathbf{1}\mathbf{1}^T-\mI-\widehat{\mG}))\|_F^2=\| \underbrace{(\mathbf{1}\mathbf{1}^T-\mI-\mG)}_{\triangleq \mD'}+\mD)\|_F^2
\end{align*}
and now we will basically decompose the loss into two orthogonal terms as follows using the centering matrix i.e. Householder transformation $\mH=\mI-\frac{1}{N}\mathbf{1}\mathbf{1}^T$ and noting that $\mH\mathbf{1}=\mathbf{0}$. We simplify
\begin{align*}
    \| \mD' -\mD\|_F^2=&\left\|(\mH+\frac{1}{N}\mathbf{1}\mathbf{1}^T) \left( \mD' -\mD\right)(\mH+\frac{1}{N}\mathbf{1}\mathbf{1}^T) \right\|_F^2\;\;\;\;\;\text{(since $\mH+\frac{1}{N}\mathbf{1}\mathbf{1}^T=\mI$)}\\
    =&\left\|\mH\left( \mD' -\mD\right)\mH \right\|_F^2
    +2\Tr\left(\mH\left( \mD' -\mD\right)\frac{1}{N}\mathbf{1}\mathbf{1}^T\right)
    +\left\|\frac{1}{N}\mathbf{1}\mathbf{1}^T\left( \mD' -\mD\right)\frac{1}{N}\mathbf{1}\mathbf{1}^T \right\|_F^2\\
    =&\left\|\mH\left( \mD' -\mD\right)\mH \right\|_F^2
    +\left\|\frac{1}{N}\mathbf{1}\mathbf{1}^T\left( \mD' -\mD\right)\frac{1}{N}\mathbf{1}\mathbf{1}^T \right\|_F^2\\
    =&\left\|\mH\left( \mD' -\mD\right)\mH \right\|_F^2
    +(\mean(\mD)-\mean(\mD'))^2\\
    =&4\left\|-\frac{1}{2}\mH\mD'\mH -(-\frac{1}{2}\mH\mD\mH) \right\|_F^2
    +(\mean(\mD)-\mean(\mD'))^2\\
    =&4\left\|-\frac{1}{2}\mH\mD'\mH -\mH\mZ\mZ^T\mH \right\|_F^2
    +(\mean(\mD)-\mean(\mD'))^2\\
    =&4\left\|\mH(\mG+\mI-2\mZ\mZ^T)\mH \right\|_F^2
    +(\mean(\mD)-\mean(\mD'))^2
\end{align*}
which can then be minimized using \citep{higham1988computing} to obtain that $\mZ$ will be proportional to the first $K$ eigenvectors of $\mG+\mI$ rescaled by the square root of their eigenvalues divided by $2\tau$. For the case with softmax and cross-entropy, we entirely leverage on Corollary 2.2 of \citet{ganea2019breaking} (a study done in the context of the softmax dimensional bottleneck \citep{yang2017breaking}) where it was shown that the minimum of the cross entropy loss is obtained whenever the rank of the pre-activations (logits) of the true and predicted probabilities have same rank. As soon as one uses an $\epsilon$ label-smoothing, to ensure that the logits of the distribution (rows of $\mG$) do not go to infinity in the case where a row contains only a single $1$, we obtain the desired result. For the RBF case, it is proven to be always nonsingular,
regardless of the value of the temperature parameter, as long as the samples
are all distinct \citep{knaf2007kernel}, hence in our case, the only solution to minimize the loss is for the samples (features maps in this case) to collapse i.e. for the rank of $\mZ$ to align with the one of $\mG$, see \citet{wathen2015spectral} for other kernels.

\subsection[Proof]{SimCLR/NNCLR recovers ISOMAP and MDS (\cref{thm:simclr_mds})}
\label{proof:simclr_mds}
This proof essentially follows the same derivations than \cref{proof:simclr} up until the last equality. Then, one can see that the eigenvectors of $\mG$ and $\mG+\mI$ are identical (only the eigenvalues are shifted by $1$). Hence, up to a rescaling, solving the ISOMAP optimization problem or the SimCLR one are equivalent, up to shift and rescaling of the representations.

\subsection[Proof]{Barlow Twins optimal Representation (\cref{thm:BT})}
\label{proof:BT}

Canonical correlation analysis usually takes the form of an optimization problem searching over linear weights $\mW$ that maximally correlates the transformed inputs. In our case, we ought to work directly with the representation as it is the final input representation that is being fed into the BarlowTwins loss. To ease this proof, we will heavily rely on the derivations from \cref{proof:BT_cca} since it offers the exact link between BarlowTwins and Canonical Correlation Analysis. The kernel version is nothing more that linear CCA but with input the $\phi(\vx_n)$ representations as opposed to the actual inputs $\vx_n$. Hence \cref{proof:BT_cca} can be used in the same way to obtain that BarlowTwins with a nonlinear DN recovers Kernel CCA. Now, the key result concerns the rank of the representation. First, we ought to recall that we are working with the regularized version of CCA that adds a small constant to the denominator of the cosine similarity computation between the columns of $\mZ_{\rm a}$ and $\mZ_{\rm b}$.

\subsection[Proof]{Linear VICReg is Locality Preserving Projection  (\cref{thm:vicreg_linear})}
\label{proof:vicreg_lpp}

For the following result, we will assume that $\mX^T(\mD-\mG)\mX$ is invertible for clarity of notations, although not required for a solution to exist \citep{liang2013trace}. This assumption holds as long as the within-connected components (resp. within-class) variance of the samples is positive. Given that, it is will known that Laplacian Eigenmap with linear mapping of the input produces LPP, hence the proof essential relies on \cref{proof:vicreg_le} that tied VICRef with Laplacian Eigenmap, and then e.g. on \citep{kokiopoulou2011trace} for that known relationship between those two models.

\subsection[Proof]{Linear VICReg with supervised relation matrix is LDA (\cref{thm:vicreg_linear})}
\label{proof:vicreg_lda}

We should highlight that the exact LDA employs an optimization problem that is typically nonconvex, and for which there does not exist a closed-form solution \citep{webb2003statistical}. Hence it is common to look at a simpler optimization problem instead \citep{guo2003generalized,wang2007trace} which is the one that VICReg closed-form solution recovers. For this section and without loss of generation we simply our notations by assuming that $\mG$ as a degree matrix of $\mI$ i.e. the sum of each row sum to one.
The objective in 2-class LDA (see \citet{li2014fisher} for examples) is to minimize the following objective
\begin{align*}
    \max\frac{\vw^T\mSigma_{b}\vw}{\vw^T\mSigma_{w}\vw},
\end{align*}
where $\mSigma_{t}=\mX^T\mX,\mSigma_{w}=\mX^T(\mI-\mG)\mX,\mSigma_{b}=\mX^T\mG\mX$ encode the total/within/between cluster variances respectively using the supervised relation matrix $\mG$. In fact, (assuming for clarity that the inputs have $0$ mean) one has
\begin{align*}
    \frac{1}{N}\mX^T\mG\mX=& \frac{1}{N}\sum_{c=1}^{C}\sum_{i \in \mathcal{N}_c}\sum_{j \in \mathcal{N}_c}\frac{1}{N_c}\vx_i\vx_j^T\\
    =& \frac{1}{N}\sum_{c=1}^{C}N_c\sum_{i \in \mathcal{N}_c}\sum_{j \in \mathcal{N}_c}\frac{1}{N_c^2}\vx_i\vx_j^T\\
    =& \frac{1}{N}\sum_{c=1}^{C}N_c\frac{\sum_{i \in \mathcal{N}_c}\vx_i}{N_c}\frac{\sum_{j \in \mathcal{N}_c}\vx_j^T}{N_c}\\
    =& \sum_{c=1}^{C}\frac{N_c}{N}\vmu_c \vmu_c^T,
\end{align*}
recovering the between (inter-cluster) variance $\mSigma_{b}$ with $\vmu_c$ the center of class $c$. Now, since we have that $\mSigma_{t}=\mSigma_{b}+\mSigma_{w}$ we directly have that $\mSigma_{w}=\frac{1}{N}\left(\mX^T\mX-\mX^T\mG\mX\right)=\frac{1}{N}\left(\mX^T\left(\mI-\mG\right)\mX\right)$. Given that, one can directly solve the LDA problem (in this case we provide directly the multivariate setting) via
\begin{align*}
    \max_{\mW}\frac{|\mW^T\mSigma_{b}\mW |}{| \mW^T\mSigma_{w}\mW|},
\end{align*}
which is known as the Fisher criterion. Whenever $\mSigma_{w}$ is invertible (if not, then the original data can be projected to a lower dimensional subspace without loss on information that would ensure that $\mSigma_{w}$ is invertible) then Fisher’s criterion is maximized leading to $\mW$ being the solution to the generalized eigenvalue problem
\begin{align*}
    \mW^*=&\argmax \frac{|\mW^T\mSigma_{b}\mW |}{| \mW^T\mSigma_{w}\mW|}=\text{top eigenvectors of: } (\mSigma_{w})^{-1}\mSigma_{b}.
\end{align*}
Notice that the generalized eigenvalue problem is exactly
\begin{align*}
    \mX^T\mG\mX\mW=\diag(\lambda)\mX^T(\mI-\mG)\mX\mW,
\end{align*}
and that $\mG \mathbf{1}=\mathbf{1}$ we observe that this is exactly the same solution than the LPP problem with $\mG$ being the relation matrix and $\mI$ being the degree matrix of the graph. Hence we obtain that LPP is equivalent to LDA whenever one uses the supervised relation matrix $\mG$ as $\mG$ (which was first pointed out in \cite{kokiopoulou2011trace}) and that it corresponds exactly to VICReg in the linear regime (as per \cref{proof:vicreg_lpp}). We also note that the eigenvalue of the above matrix $(\mSigma_{w})^{-1}\mSigma_{b}$ might seem to be arbitrary. However we have the following lemma that ensure that the eigenvalues of are real nonnegative.

\begin{lemma}
\label{lemma:NN}
The eigenvalues of $(\mSigma_{w})^{-1}\mSigma_{b}$ are equal to the eigenvalues of $\mSigma_{w}^{-1/2}\mSigma_{b}\mSigma_{w}^{-1/2}$ which is symmetric hence all are nonegative and real.
\end{lemma}

\subsection[Proof]{Proof of BarlowTwins in the Linear Regime (\cref{thm:BT_cca})}
\label{proof:BT_cca}

The tie between CCA and LDA is not new. In fact, if one considers one dataset to be the samples and the other to be the class labels (binary variables for two-class problems or a variation of one-hot encoding for multi-class problems) then CCA and LDA are equivalent \citep{bartlett1938further,hastie1995penalized}. But in our case, we do not explicitly use the labels but the views of different class. Hence, the goal of this section is to first recover the analytical parameters of BarlowTwins in the linear regime through CCA and then to demonstrate that the CCA-LDA connection also persists in the case when the views are both obtained from per-class samples i.e. exactly fitting the BarlowTwins scenario.

There exists many different ways to formulate the CCA problem. At the most simple level, one aims to sequentially learn pairs of filters that produce maximally correlated features as in
\begin{align*}
    \max_{\vw_{\rm a},\vw_{\rm b}}\frac{\langle \mX_{\rm a}\mW_{\rm a} ,\mX_{\rm b}\mW_{\rm b} \rangle }{\|\mX_{\rm a}\mW_{\rm a} \|_2 \| \mX_{\rm b}\mW_{\rm b}\|_2},
\end{align*}
which can easily be recognized to be the diagonal element of the BarlowTwins loss that we aim to maximize in the linear regime i.e. the cosine similarity between a column of $\mZ_{\rm a}$ and the same column of $\mZ_{\rm b}$. To find the corresponding filters, which are in general not assume to be identical, one transforms the above problem to a constrained optimization problem exactly as done with eigenvalue problems with the Rayleigh quotient. We thus obtain the following
\begin{align*}
    \max_{\vw_{\rm a},\vw_{\rm b}} \langle \mX_{\rm a}\mW_{\rm a} ,\mX_{\rm b}\mW_{\rm b} \rangle \text{ s.t. }\|\mX_{\rm a}\mW_{\rm a} \|_2=1 \text{ and } \| \mX_{\rm b}\mW_{\rm b}\|_2=1,
\end{align*}
since rescaling of the weights does not impact the learned filters, which can be transformed into a  Lagrangian function as
\begin{align*}
    \mathcal{L} = \langle \mX_{\rm a}\mW_{\rm a} ,\mX_{\rm b}\mW_{\rm b} \rangle -\lambda_1(\|\mX_{\rm a}\mW_{\rm a} \|_2-1) -\lambda_2(\| \mX_{\rm b}\mW_{\rm b}\|_2-1),
\end{align*}
which has the following partial derivatives using the $\mSigma_{\rm ab}$ notations for the cross-products
\begin{align*}
    \nabla_{\vw_{\rm a}} \mathcal{L} =& \mSigma_{\rm ab}\vw_{\rm b}-2\lambda_1 \mSigma_{\rm aa} \vw_{\rm a}\\
    \nabla_{\vw_{\rm b}} \mathcal{L} =& \mSigma_{\rm ba}\vw_{\rm a}-2\lambda_2 \mSigma_{\rm bb} \vw_{\rm b}\\
    \frac{\partial \mathcal{L}}{\partial \lambda_1} =&\vw_{\rm a}^T \mSigma_{\rm aa}\vw_{\rm a}\\
    \frac{\partial \mathcal{L}}{\partial \lambda_2} =&\vw_{\rm b}^T \mSigma_{\rm bb}\vw_{\rm b},
\end{align*}
the first thing to notice is that setting those (first two) to zero will lead to $\lambda_1=\lambda_2$ which we thus denote as a single $\lambda$ parameter since 
\begin{align*}
    \mSigma_{\rm ab}\vw_{\rm b}-2\lambda_1 \mSigma_{\rm aa} \vw_{\rm a}=0\implies \vw_{\rm a}^T\mSigma_{\rm ab}\vw_{\rm b}-2\lambda_1 \vw_{\rm a}^T\mSigma_{\rm aa} \vw_{\rm a}=0\implies \lambda_1 = \frac{1}{2}\vw_{\rm a}^T\mSigma_{\rm ab}\vw_{\rm b}\\
    \mSigma_{\rm ba}\vw_{\rm a}-2\lambda_2 \mSigma_{\rm bb} \vw_{\rm b}=0\implies \vw_{\rm b}^T\mSigma_{\rm ba}\vw_{\rm a}-2\lambda_2 \vw_{\rm b}^T\mSigma_{\rm bb} \vw_{\rm b}=0\implies \lambda_2 = \frac{1}{2}\vw_{\rm b}^T\mSigma_{\rm ba}\vw_{\rm a},
\end{align*}
and thus we can simplify the original system using a single Lagrangian multiplier to have th following system
\begin{align*}
    \begin{array}{r}
    \mSigma_{\rm ab}\vw_{\rm b}-2\lambda \mSigma_{\rm aa} \vw_{\rm a}=0\\
    \mSigma_{\rm ba}\vw_{\rm a}-2\lambda \mSigma_{\rm bb} \vw_{\rm b}=0
    \end{array}\Big \} 
    \begin{array}{r}
    \vw_{\rm a}=\frac{1}{2\lambda }\mSigma_{\rm aa}^{-1}\mSigma_{\rm ab}\vw_{\rm b}\\
    (\frac{1}{4}\mSigma_{\rm bb}^{-1}\mSigma_{\rm ba}\mSigma_{\rm aa}^{-1}\mSigma_{\rm ab}-\lambda^2 \mI)\vw_{\rm b}=\mathbf{0}
    \end{array}\Big \},
\end{align*}
where it is clear that one obtains $\vw_{\rm b}$ by solving the generalized eigenvalue problem and then finds $\vw_{\rm a}$ accordingly and this can be done by remove the $\frac{1}{4}$ factor and replacing $2\lambda$ by $\lambda$. For specific implementations of the above, we direct the reader to \citet{healy1957rotation,ewerbring1989canonical,hardoon2004canonical}.

{\bf Recovering LDA from supervised CCA.}~To recover this result, we closely follow the methodology from \citet{kursun2011canonical}. 
Additionally, we obtain that in this case the two views are permutation of each other due to the symmetric structure of $\mG$ that simply shuffles around the (repeated) samples to consider all the within-class pairs. Recall that even in t his case $\mG$ as for degree matrix $\mI$ since we only use positive pairs between the samples (the dataset has been augmented first duplicating each input in each class to match with all others). In that setting, we first have the following simplifications
We obtain \begin{align*}
    &\mSigma_{aa}=\mZ^T\mZ,\;\;
    \mSigma_{ab}=\mZ^T\mG\mZ,\;\;
    \mSigma_{ba}=\mZ^T\mG\mZ,\;\;
    \mSigma_{bb}=\mZ^T\mZ,
\end{align*}
since we have that $\mG=\mG^T$ and $\mG^2=\mI$. Using the result from 
CCA \cite{uurtio2017tutorial} we have the the CCA is equal to the sum of the singular values of $\mSigma_{bb}^{-1}\mSigma_{ba}C^{-1}_{aa}\mSigma_{ab}$ which corresponds to $\Tr((\mZ^T\mZ)^{-1}\mZ^T\mG\mZ)$ since $\mSigma_{ab}=\mSigma_{ba}$ and $\mSigma_{aa}=\mSigma_{bb}$.
First, notice that the CCA optimum solves the following eigenvalue problem
\begin{align*}
    \mSigma_{aa}^{-1}\mSigma_{ab}\mSigma_{bb}^{-1}\mSigma_{ba}\vu_{\rm CCA}=\lambda_{\rm CCA} \vu_{\rm CCA},
\end{align*}
but noticing that in our case $\mX_{\rm l}$ and $\mX_{\rm r}$ are permutations of each other, we directly obtain that
\begin{align*}
    \mSigma_{aa}^{-1}\mSigma_{ab}\mSigma_{bb}^{-1}\mSigma_{ba}\vu_{\rm CCA}=\lambda_{\rm CCA} \vu_{\rm CCA}
    \iff\mSigma_{t}^{-1}\mSigma_{b}\mSigma_{t}^{-1}\mSigma_{b}\vu_{\rm CCA}=\lambda_{\rm CCA} \vu_{\rm CCA},
\end{align*}
using the total, between and within-class covariance matrices. Now, starting from the LDA loss, we will see that we recover the CCA one thanks to the above facts
\begin{align*}
    \mSigma_{b}\vu_{\rm LDA}=&\lambda_{\rm LDA}\mSigma_{w} \vu_{\rm LDA} &&\text{(original LDA parameter)}\\
    \iff\mSigma_{b}\vu_{\rm LDA}=&\lambda_{\rm LDA}(\mSigma_{t}-\mSigma_{b}) \vu_{\rm LDA}\\
    \iff\mSigma_{b}(1+\lambda_{\rm LDA})\vu_{\rm LDA}=&\lambda_{\rm LDA}\mSigma_{t} \vu_{\rm LDA}\\
    \iff\mSigma_{b}\vu_{\rm LDA}=&\frac{\lambda_{\rm LDA}}{(1+\lambda_{\rm LDA})}\mSigma_{t} \vu_{\rm LDA}\\
    \iff \mSigma_{t}^{-1}\mSigma_{b}\vu_{\rm LDA}=&\frac{\lambda_{\rm LDA}}{(1+\lambda_{\rm LDA})} \vu_{\rm LDA}\\
    \iff \mSigma_{t}^{-1}\mSigma_{b}\mSigma_{t}^{-1}\mSigma_{b}\vu_{\rm LDA}=&\left(\frac{\lambda_{\rm LDA}}{1+\lambda_{\rm LDA}}\right)^2 \vu_{\rm LDA}&&\text{(original CCA problem)}
\end{align*}
hence given the LDA or CCA solutions one can recover the others, the only difference lies in the singular values, which can also be found from each other. Also, recall that the eigenvalues are real positive as per \cref{lemma:NN} ensuring the well posed division used above.

\subsection[Proof]{Proof of VICReg Optimality (\cref{thm:optimal})}
\label{proof:optimal_vicreg}

The proof first uses \cref{thm:VICReg_optimal} that demonstrates how the optimal representation $\mZ$ will have the left singular vectors of $\mG$ up to rotations for the ones with multiplicity greater than one. Then, using that, we obtain from \cref{thm:optimal} that if those singular vectors are able to span the left singular vectors of $\mY$ then there exists a linear probe that will solve the task at hand, giving the desired result.





\end{document}